\documentclass[twoside]{article}

\usepackage[accepted]{aistats2016}
\usepackage[breaklinks]{hyperref}
\usepackage[hyphenbreaks]{breakurl}

\usepackage{url}
 \usepackage{etoolbox}

\usepackage{amssymb}
\usepackage{natbib}

\usepackage{graphicx}
\bibpunct{(}{)}{;}{a}{,}{,}

\usepackage{amsmath,amssymb,graphicx,bbm,enumerate}
\usepackage{verbatim}
\usepackage{bbm}
\graphicspath{{figures/}}
\usepackage{color}
\usepackage{dsfont}
\usepackage{sectsty}
\definecolor{dukeblue}{RGB}{0, 0, 156}

\usepackage{adjustbox}

\usepackage{subfigure}
\usepackage{multirow}

\usepackage{moresize}
\usepackage[flushleft]{threeparttable}
\usepackage{multirow}
\usepackage{array}
\usepackage{natbib}
\newcolumntype{I}{!{\vrule width 1pt}}
\newlength\savedwidth

\usepackage{enumitem}
\usepackage[font=small]{caption}
\usepackage{algorithm}
\usepackage{algorithmic}

\def\cN{{\mathcal N}} 
\def\diag{{\rm diag}}

\def\cG{{\mathcal G}} 

\def\wt#1{\widetilde{#1}} 
\def\bd#1{\boldsymbol{#1}}
\mathchardef\mhyphen="2D 

\def\argmin{\operatornamewithlimits{arg\,min}}

\newcommand{\rmnum}[1]{\textit{\romannumeral #1}}

\def\beginmat{ \left( \begin{array} }
\def\endmat{ \end{array} \right) }
\def\diag{{\rm diag}}
\def\log{{\rm log}}

\graphicspath{{figure/}}

\begin{document}

%
\runningtitle{Variational Gaussian Copula Inference }

%
\runningauthor{Shaobo Han, Xuejun Liao,  David B. Dunson, Lawrence Carin}

\twocolumn[

\aistatstitle{Variational  Gaussian Copula Inference}

  \aistatsauthor{ Shaobo Han \And  Xuejun Liao \And David B. Dunson$^{\dag}$ \And Lawrence Carin }

   \aistatsaddress{ Department of ECE,  Department of Statistical Science$^{\dag}$,   Duke University,  Durham, NC 27708, USA } ]




\begin{abstract}

We utilize copulas to constitute a unified framework for constructing and optimizing
variational proposals in hierarchical Bayesian models. For models with continuous and non-Gaussian hidden variables, we propose a {semiparametric} and {automated} variational  Gaussian copula approach, in which the parametric Gaussian copula family is able to preserve multivariate posterior dependence, and the nonparametric transformations based on Bernstein polynomials provide ample flexibility in characterizing the univariate marginal posteriors.

\end{abstract}

\section{Introduction}

A crucial component of Bayesian inference is approximating the posterior distribution, which represents the current state of knowledge about the latent  variables  $\bd{x}$ after data  $\bd{y}$  have been observed. When intractable integrals are involved, variational inference methods find an approximation $q(\bd{x})$ to the posterior distribution $p(\bd{x}|\bd{y})$ by minimizing the Kullback-Leibler (KL) divergence
$\mathrm{KL}\{q(\bd{x})||p(\bd{x}|\bd{y})\}=\int q(\bd{x})\log\left[{{q(\bd{x})}/{p(\bd{x}|\bd{y})}}\right]d\bd{x}$,   providing a lower bound for the marginal likelihood. 

To make inference tractable,  mean-field variational Bayes (MFVB) methods \citep{jordan1999introduction, wainwright2008graphical} assume $q(\bd{x})$ is factorized over a certain partition of the latent variables $\bd{x}\equiv[\bd{x}_{1}, \hdots, \bd{x}_{J}]$, $q_{\mathrm{VB}}(\bd{x})=\prod_{j}q_{\mathrm{VB}}(\bd{x}_{j})$, with marginal densities $q_{\mathrm{VB}}(\bd{x}_{j})$ in free-form and correlations between partitions neglected. The structured mean-field approaches \citep{saul1996exploiting, hoffman2014structured} preserve partial correlations and apply only to models with readily identified substructures. The variational Gaussian (VG) approximation \citep{barber1998ensemble, opper2009variational} allows incorporation of correlations by postulating a multivariate Gaussian parametric form $q_{\mathrm{VG}}(\bd{x})=\cN(\bd{\mu}, \bd{\Sigma})$. The VG approximation, with  continuous margins of real variables, are not suitable for  variables that are inherently positive or constrained, skewed, or heavy tailed.  For multi-modal posteriors, a mixture of MFVB \citep{VB-1998-MixtureMeanField} or a mixture of uniformly-weighted Gaussians \citep{GershmanBlei} may be employed, which usually requires a further lower bound on the average over the logarithm of the mixture distribution. 

To address the limitations of current variational methods in failing to simultaneously characterize the posterior dependencies among latent variables while  allowing skewness, multimodality, and other characteristics, we propose a new variational copula framework.  Our approach decouples the overall inference task into two subtasks: (\textit{i}) inference of the copula function, which captures the multivariate posterior dependencies; (\textit{ii}) inference of a set of univariate margins, which are allowed to take essentially any form. Motivated by the work  on automated (black-box) variational inference \citep{ranganath2013black,mnih2014neural, titsias2014doubly, nguyen2014automated, Dkingma2014}, we  present a  stochastic optimization algorithm  for \emph{generic} hierarchical Bayesian models with continuous variables, which (\rmnum{1}) requires minimal model-specific derivations,  (\rmnum{2}) reproduces  peculiarities of the true marginal posteriors, and (\rmnum{3}) identifies interpretable  dependency structure among latent variables. 

Using copulas to improve approximate Bayesian inference is a natural idea that has also been  explored recently in other contexts \citep{li2015extending,ferkingstad2015improving}. Independently from our work, \cite{Dtran2015}  presented a  copula augmented variational method with fixed-form marginals, and utilizes regular vines to decompose the multivariate dependency structure into bivariate copulas and a nest of trees. Our method  provides complementary perspectives on nonparametric treatment of univariate marginals. 

 \section{Variational Copula Inference Framework}

 Sklar's theorem \citep{1959fonctions} ensures that  any multivariate joint distribution $Q$ can be written in terms of univariate marginal distributions $F_{j}(x)=P(X_{j}\leq x)$, $j=1, \hdots, p$ and a copula which describes the dependence structures between variables, such that 
 \begin{align}\label{eq1}
 Q(x_{1}, \hdots, x_{p})=C[F_{1}(x_{1}),\hdots, F_{p}(x_{p})].
 \end{align}
 Conversely, if $C$ is a copula and $\{F_{j}\}_{j=1:p}$ are distribution functions, then the function $Q$ defined by \eqref{eq1}  is a $p\mhyphen$dimensional joint distribution function with marginal distributions $F_{1}, F_{2}, \hdots, F_{p}$, owing to the marginally closed property \citep{xue2000multivariate}. Assuming $Q(x_{1}, ..., x_{p})$ has $p$-order partial derivatives, the joint probability density function (PDF) is 
$q(x_{1}, \hdots, x_{p}) = c_{\bd{\Theta}}[F_{1}(x_{1}),\hdots, F_{p}(x_{p}) ]\prod_{j=1}^{p}f_{j}(x_{j})$, where $f_{j}(x_{j})$ is the PDF of the $j$th variable and it is related to the corresponding cumulative distribution function (CDF) by $F_{j}(x_{j})=\int_{-\infty}^{x}f_{j}(t) d t$, $c_{\bd{\Theta}}$ is the copula density with parameter $\bd{\Theta}$. 

 Sklar's theorem allows separation of the marginal distributions $F_{j}(x_{j})$ from the dependence structure, which is appropriately expressed in the copula function $C$.  As a modeling tool, the specified copula function and margins  can be directly fitted to the observed data $\bd{y}$ \citep{liu2009nonparanormal, wauthier2010heavy, lopez2013gaussian} 
 with their parameters optimized via Bayesian or maximum likelihood estimators (see \cite{smith2013bayesian} and the references therein). 
 In contrast, our goal is to use a copula as an \emph{inference engine} for full posterior approximation.  
  All the unknowns (variables/parameters) in the user-specified hierarchical model are encapsulated into a vector $\bd{x}$, and  the optimal variational approximation $q_{\mathrm{VC}}(\bd{x})$ to the true posterior $p(\bd{x}|\bd{y})$ is found under the Sklar's representation. This approach provides users with full modeling freedom and does not require conditional conjugacy between latent variables; thus the approach is applicable to general models.  
Within some tractable copula family $C\in \mathcal{C}$,  and assuming $F(\cdot)$ and $C(\cdot)$ to be differentiable, we construct the variational proposal  as
$q_{\mathrm{C}}(\bd{x})= c(\bd{u}) \prod_{j=1}^{p}f_{j}(x_{j})$, where $\bd{u}=F(\bd{x})=[F_{1}(x_{1}), \hdots, F_{p}(x_{p})]$,  such that the approximation  satisfies 
\begin{align}
 q_{\mathrm{C}}^{\star}(\bd{x})&=\argmin_{q_{\mathrm{C}}(\bd{x})} \mathrm{KL}\{q_{\mathrm{C}}(\bd{x})||p(\bd{x}|\bd{y})\}\cr 
 &=\argmin_{q_{\mathrm{C}}(\bd{x})}  \mathrm{KL}\{q_{\mathrm{C}}(\bd{x})||p(\bd{x})\}-\mathbb{E}_{q_{\mathrm{C}}(\bd{x})}[\ln {p(\bd{y}|\bd{x})}],\nonumber
\end{align}
where $p(\bd{y}|\bd{x})$ is the likelihood and $p(\bd{x})$ is the prior. Letting the true posterior $p(\bd{x}|\bd{y})$ in Sklar's representation be  $p(\bd{x}|\bd{y}) =c^{\star}(\bd{v})\prod_{j}f_{j}^{\star}(x_{j})$,  where $\bd{v}=[F_{1}^{\star}(x_{1}), \hdots, F_{p}^{\star}(x_{p})]$, $c^{\star}(\bd{v})$ and $\{f_{j}^{\star}(x_{j})\}_{j=1:p}$ are the true underlying copula density and marginal posterior densities, respectively, the KL divergence decomposes into additive terms (derivations are provided in Supplementary Material), 
\begin{align}\label{eq2}
\mathrm{KL}\{q_{\mathrm{C}}(\bd{x})||p(\bd{x}|\bd{y})\} &=\mathrm{KL}\{c[F(\bd{x})]||c^{\star}[F^{\star}(\bd{x})]\} \nonumber\\ & +\sum\nolimits_{j}\mathrm{KL}\{f_{j}(x_{j})||f_{j}^{\star}(x_{j})\}.
\end{align}
Classical methods, such as MFVB and the VG approximation are  special cases of the proposed VC  inference framework. We next compare their KL divergence under Sklar's representation and offer a reinterpretation of them under the proposed framework.

\subsection{Special Case 1:  Mean-field VB}

The mean-field proposal  corresponds to the independence copula $C_{\Pi}(\bd{u})=\prod_{j=1}^{J}u_{j}$ with free-form marginal densities $f_{j}(\bd{x}_{j})$. Given  $c_{\Pi}(\bd{u})= 1$  we have  $q_{\Pi}(\bd{x})=c_{\Pi}(\bd{u})\prod_{j}f_{j}(\bd{x}_{j})=\prod_{j}f_{j}(\bd{x}_{j})=q_{\mathrm{VB}}(\bd{x})$. If MFVB is not fully factorized, i.e. $J<p$, the independence copula is the only copula satisfying the marginal closed property, according to the impossibility theorem \citep{nelsen2007introduction}.
 MFVB assumes an independence copula and only optimizes the free-form margins, 
\begin{align}\label{eq3}
\mathrm{KL}\{q_{\mathrm{VB}}(\bd{x})||p(\bd{x}|\bd{y})\} &=\mathrm{KL}\{c_{\Pi}[F(\bd{x})]||c^{\star}[F^{\star}(\bd{x})]\} \nonumber\\ &+\sum\nolimits_{j}{\mathrm{KL}\{f_{j}(x_{j})||f_{j}^{\star}(x_{j})\}}.
\end{align}
The lowest achievable KL divergence in  MFVB is $\mathrm{KL}\{q_{\mathrm{VB}}(\bd{x})||p(\bd{x}|\bd{y})\}=\mathrm{KL}\{c_{\mathrm{\Pi}}[F(\bd{x})]||c^{\star}(F(\bd{x}))\}$, which is achieved when the true posterior marginals are found, i.e. $F_{j}\equiv F^{\star}_{j}, \forall j$ , in which case the overall KL divergence is reduced to the KL divergence  between the independence copula and the true copula. As is shown in \eqref{eq3}, the objective function contains two terms,  both involving  marginal CDFs $\{F_{j}\}_{j=1:p}$. Since in general $c^{\star}\neq{}c_{\Pi}$, the optimal $F$ minimizing the first term will not be equal to $F^\star$. Therefore, minimizing \eqref{eq3} will not lead to the correct marginals and this partially explains the reason why MFVB usually cannot find the true marginal posteriors in practice (e.g., variances can be severely underestimated \citep{neville2014mean}), even though it allows for free-form margins.

\subsection{Special Case 2: VG Approximation}

In fixed-form variational Bayes \citep{honkela2010approximate}, such as VG approximation, the multivariate Gaussian proposal $q_{\mathrm{VG}}(\bd{x})=\cN(\bd{x}; \bd{\mu}, \bd{\Sigma})$ can be written as $q_{\mathrm{VG}}(\bd{x})=c_{\mathrm{G}}(\bd{u}|\bd{\Upsilon})\prod_{j=1}^{p}\phi_{j}(x_{j}; \mu_{j}, \sigma_{j}^{2})$. VG not only assumes the true copula function is a Gaussian copula \citep{xue2000multivariate} with parameter $\bd{\Upsilon}=\bd{D}^{-{1}/{2}}\bd{\Sigma}\bd{D}^{-{1}/{2}}$, $\bd{D}=\diag(\bd{\Sigma})$, 
but is also restricted to univariate Gaussian  marginal densities $\{\phi_{j}(x_{j}; \mu_{j}, \sigma_{j}^{2})\}_{j=1:p}$,  
\begin{align}\label{eq4}
\mathrm{KL}\{q_{\mathrm{VG}}(\bd{x})||p(\bd{x}|\bd{y})\}& =\mathrm{KL}\{c_{\mathrm{G}}[\Phi(\bd{x})]||c^{\star}[F^{\star}(\bd{x})]\}\nonumber \\ &+\sum\nolimits_{j}{\mathrm{KL}\{\phi_{j}(x_{j})||f_{j}^{\star}(x_{j})\}}.
\end{align}
We can see in \eqref{eq4} that if the margins are misspecified,  even if the true underlying copula is a Gaussian copula, $c_{\mathrm{G}}\equiv c^{\star}$,
there could still be a discrepancy $\sum_{j}{\mathrm{KL}\{\phi_{j}(x_{j})||f_{j}^{\star}(x_{j})\}}$ between margins, and $\mathrm{KL}\{c_{\mathrm{G}}[\Phi(\bd{x})]||c^{\star}[F^{\star}(\bd{x})]\}$ is  not zero. 

Concerning analytical tractability and simplicity,  in the sequel we  concentrate  on variational Gaussian copula (VGC) proposals constructed via Gaussian copula with continuous margins, i.e. $q_{\mathrm{VGC}}(\bd{x}) = c_{\mathrm{G}}(\bd{u}|\bd{\Upsilon})\prod_{j=1}^{p}f_{j}(x_{j})$,
 where $\bd{u}=[F_{1}(x_{1}), \hdots, F_{p}(x_{p})]$.
Our VGC method extends MFVB and VG,  and  improves upon both  by allowing simultaneous  updates of  the Gaussian copula parameter  $\bd{\Upsilon}$ and the adaptation of marginal densities $\{f_{j}(x_{j})\}_{j=1:p}$. First, the univariate margins in VGC is not restricted to be Gaussian. Second, the Gaussian copula in VGC is more resistant to local optima than the independence copula assumed in MFVB and alleviates its variance underestimation pitfall, as is demonstrated in Section \ref{secNIGG}. 



\section{Variational Gaussian Copula Approximation}

A Gaussian copula function with $p\times p$ correlation matrix $\bd{\Upsilon}$ is defined as 
$C_{\mathrm{G}}(u_{1}, \hdots, u_{p}|\bd{\Upsilon})=\Phi_{p}(\Phi^{-1}(u_{1}), \hdots,\Phi^{-1}(u_{p})|\bd{\Upsilon}):  [0,1]^{p}\rightarrow [0,1]$ where $\Phi(\cdot)$ is a shorthand notation of the CDF of  $\cN(0, 1)$, and $\Phi_{p}(\cdot|\bd{\Upsilon})$ is the CDF of $N_{p}(\bd{0}, \bd{\Upsilon})$.  The Gaussian copula density is
\begin{align}
c_{\mathrm{G}}(u_{1}, \hdots, u_{p}|\bd{\Upsilon})& = \frac{1}{\sqrt{|\bd{\Upsilon}|}}{\exp{\left\{-\frac{\bd{z}^{T}( \bd{\Upsilon}^{-1}-\bd{I}_{p})\bd{z}}{2}\right\}}}, \nonumber
\end{align}
where $\bd{z} =[\Phi^{-1}(u_{1}), \hdots, \Phi^{-1}(u_{p})]^{T}$. 

In the proposed VGC approximation,  the  variational proposal $q_{\mathrm{VGC}}(\bd{x})$ is constructed  as a product of Gaussian copula density and continuous marginal densities. 
The evidence lower bound (ELBO) of VGC approximation is 
\begin{align}\label{eq5}
\mathcal{L}_{\mathrm{C}}[q_{\mathrm{VGC}}(\bd{x})]&=\int \bigg[c_{\mathrm{G}}[F(\bd{x})]\times \prod_{j=1}^{p}f_{j}(x_{j})\bigg]\ln{{p(\bd{y},\bd{x})}}d\bd{x} \nonumber\\ & + H[c_{\mathrm{G}}(\bd{u})]+\sum_{j=1}^{p} H[{f_{j}}(x_{j})],
\end{align}
where ${u}_{j}=F_{j}(x_{j})$, $H[f(x)]=-\int f(x)\ln{f(x)} d x$.

However, directly optimizing the ELBO in \eqref{eq5} w.r.t. the Gaussian copula parameter $\bd{\Upsilon}$ and the univariate marginals $\{f_{j}(x_{j})\}_{j=1:p}$ often leads to a non-trivial variational calculus problem.  For computational convenience, we present several equivalent proposal constructions based on Jacobian transformation and reparameterization.


\begin{table*}[bpht]
\centering
\caption{Equivalent Representations of Variational Gaussian Copula (VGC) Proposals}
\label{my-label}
\small{
\begin{tabular}{|l|l|c|c|}
\hline
 &  Posterior    Formulation &  \multicolumn{2}{c|}{Optimization Space}\\ \hline
 R0  &Original   & \multicolumn{2}{c|}{Multivariate (non-Gaussian) density $q(\bd{x})$} \\ \hline
 R1  & Sklar's Representation  &     Copula  density $c_{\mathrm{G}}(\bd{u}|\bd{\Upsilon})$      &  Univariate   marginals $\{f_{j}(x_{j})\}_{j=1:p}$         \\ \hline
 R2  & Jacobian Transform &      Gaussian density $q(\wt{\bd{z}})=\cN(\bd{0}, \bd{\Upsilon})$      & Monotone functions $\{g_{j}(z_{j})\}_{j=1:p}$           \\ \hline
 R3  & Parameter Expansion &     Gaussian density $q(\wt{\bd{z}})=\cN(\bd{\mu}, \bd{C}\bd{C}^{T})$      &      Monotone functions $\{h_{j}(\wt{z_{j}})\}_{j=1:p}$     \\ \hline
\end{tabular}}
\end{table*}
\vspace{-0.2cm}
\subsection{Equivalent Variational Proposals }
We incorporate auxiliary variables $\bd{z}$ by exploiting the latent variable representation of the Gaussian copula: $x_{j} =F_{j}^{-1}(u_{j})$, $u_{j}=\Phi({z}_{j})$, $\bd{z}\sim N_{p}(\bd{0}, \bd{\Upsilon})$. 
Letting $g_{j}(\cdot)=F_{j}^{-1}(\Phi(\cdot))$  be bijective  monotonic non-decreasing  functions, $x_{j}=g_{j}(z_{j})$, $\forall j$,  the Jacobian transformation gives 
\vspace{-0.0cm}
\begin{align}
q_{\mathrm{VGC}}(\bd{x}) &=\int \bigg[\prod_{j=1}^{p}\delta(x_{j}-g_{j}(z_{j}))\bigg]q_{\mathrm{G}}(\bd{z}; \bd{0}, \bd{\Upsilon}) d\bd{z}\cr &= q_{\mathrm{G}}(g^{-1}(\bd{x}); \bd{0}, \bd{\Upsilon})\bigg[\prod_{j=1}^{p}\frac{d}{dx_{j}}g_{j}^{-1}(x_{j})\bigg],\nonumber
\end{align}
where $\delta(\cdot)$ is the Dirac delta function. 

It is inconvenient to directly optimize the correlation matrix $\bd{\Upsilon}$ of interest, since $\bd{\Upsilon}$ is a  positive semi-definite matrix with ones on the diagonal and off-diagonal elements between $[-1,1]$. We adopt the parameter expansion (PX) technique \citep{liu1998parameter, liu1999parameter}, which has been applied in  accelerating variational Bayes \citep{jaakkola2006parameter} and {the} sampling {of} correlation matrix \citep{talhouk2012efficient}. Further considering 
$\wt{z_{j}}=t_{j}^{-1}(z_{j})=\mu_{j}+\sigma_{jj}z_{j}$,  $\wt{\bd{z}}\sim N_{p}(\bd{\mu}, \bd{\Sigma})$, $\bd{\Sigma}=\bd{D}\bd{\Upsilon}\bd{D}^{T}$,  $\bd{D}=[\diag(\sigma_{jj})]_{j=1:p}$, thus $x_{j}=g(z_{j})=g(t(\wt{z_{j}})):=h(\wt{z_{j}})$, 
where $h_{j}(\cdot)=g_{j}\circ t_{j}(\cdot)$ are also bijective  monotonic non-decreasing  functions, the variational proposal is further written as  
\begin{align}
q_{\mathrm{VGC}}(\bd{x})&=\int \bigg[\prod_{j=1}^{p}\delta(x_{j}-h_{j}(\wt{z_{j}}))\bigg]q_{\mathrm{G}}(\wt{\bd{z}}; \bd{\mu}, \bd{\Sigma}) d\wt{\bd{z}}\cr &=  q_{\mathrm{G}}(h^{-1}(\bd{x}); \bd{\mu}, \bd{\Sigma})\bigg[\prod_{j=1}^{p}\frac{d}{dx_{j}}h_{j}^{-1}(x_{j})\bigg].\nonumber
\end{align}
Given the transformations $\{h_{j}\}_{j=1:p}$, $q_{\mathrm{G}}(\wt{\bd{z}}; \bd{\mu}, \bd{\Sigma})$ can be further reparameterized by the Cholesky decomposition $\bd{\Sigma}=\bd{C}\bd{C}^{T}$\citep{challis2013gaussian, titsias2014doubly}, where $\bd{C}$ is a square lower triangular matrix. Table \ref{my-label} summarizes four translatable representations  of variational proposals. 

\subsection{VGC with Fixed-form Margins}

The ELBO under  Sklar's representation \eqref{eq5} is therefore translated into the Jacobian representation 
\vspace{-0.3cm}
\begin{align}\label{eqelbols}
& \mathcal{L}_{\mathrm{C}}[q_{\mathrm{VGC}}(\bd{x})]=\mathbb{E}_{\cN(\wt{\bd{z}}; \bd{\mu}, \bd{\Sigma})}[ \ell_{s}(\wt{\bd{z}})-\ln{q_{\mathrm{G}}(\wt{\bd{z}})}
],\cr & \ell_{s}(\wt{\bd{z}}, h) = \ln{p(\bd{y},h(\wt{\bd{z}}))}+\sum_{j=1}^{p}  \ln h_{j}'(\wt{z_{j}}).\end{align}
The monotonic transformations $h_{j}(\cdot)=F_{j}^{-1}[\Phi(t(\cdot))]$ can be specified according to the desired parametric form of marginal posterior,  if the inverse CDF $F_{j}^{-1}$ is tractable. For example, the multivariate log-normal posterior  can be constructed via a Gaussian copula with log-normal (LN) margins, 
\begin{align}
q_{\textrm{VGC-LN}}(\bd{x})=C_{\mathrm{G}}(\bd{u}|\bd{\Upsilon})\prod_{j=1}^{p}\mathrm{LN}(x_{j}; \mu_{j},\sigma_{j}^{2}).
\end{align}
This also corresponds to imposing exponential transform on Gaussian variables, $\bd{x}=h(\wt{\bd{z}})=\exp(\wt{\bd{z}})$, $\wt{\bd{z}}\sim \cN(\bd{\mu}, \bd{\Sigma})$. In this case, $\{\mu_{j},\sigma_{j}^{2}\}_{j=1:p}$ controls the location and dispersion of the marginal density;  $h(\cdot)$  does not have any additional parameters to control the shape  and $\ln h'(\wt{z_{j}})=\wt{z_{j}}$ takes a simple form. VGC-LN is further discussed in Section \ref{sec2dLN} and Section \ref{secNIGG}.

Given the copula function $C$, we only need to find $p$ one-dimensional margins. However,  without knowing characteristics of the latent variables,  specifying appropriate parametric form  for margins is a difficult task in general cases. First, the marginals might exhibit multi-modality, high skewness or kurtosis,  which are troublesome for particular parametric marginals to capture. Second, a tractable inverse CDF with optimizable arguments/parameters, as required here, are available only in a handful of cases.  Instead of using some arbitrary parametric form, we construct bijective transform functions via kernel mixtures, which lead to  highly flexible (ideally free-form) marginal proposals.

\section{Bernstein Polynomials based Monotone Transformations }

The marginal densities in VGC can be  recovered through Jacobian transformation, 
\begin{align}
f_{j}(x_{j})&=q_{\mathrm{G}}(h_{j}^{-1}({x}_{j}); \mu_{j},\sigma_{2}^{2})\frac{d}{dx_{j}}h_{j}^{-1}(x_{j})\cr &=q_{\mathrm{G}}(h_{j}^{-1}({x}_{j}); \mu_{j},\sigma_{2}^{2})\frac{1}{h_{j}'(h_{j}^{-1}(x_{j}))},
\end{align}
where the ${[h_{j}'(h_{j}^{-1}(x_{j}))]^{-1}}$ term is interpreted as a marginal-correction term. {To guarantee analytical tractability}, we require $h(\cdot)$ to be (\rmnum{1}) bijective; (\rmnum{2}) monotonic non-decreasing;  (\rmnum{3}) having unbounded/constrained range; (\rmnum{4}) differentiable with respect to both its argument and parameters;  and (\rmnum{5}) sufficiently flexible. We propose a class of continuous and smooth transformations $h(\cdot)$ constructed via kernel mixtures that automatically have these desirable properties.

\subsection{Continuous Margins Constructed via Bernstein Polynomials}

The Bernstein polynomials (BPs) have {a uniform convergence property for continuous functions on unit interval $[0,1]$} and have been used for nonparametric density estimation \citep{petrone1999bayesian}. It seems more natural to use kernel mixtures directly as the variational proposal.  However, the difficulty lies in tackling the term $f(F^{-1}(\cdot))$ involving the inverse CDF of mixtures (not analytical) and the need of a further lower bound on the entropy of mixtures.  In this paper, we overcome this issue by using a sandwich-type construction of the transform  $h(\wt{z})$\footnote{The index $j$ on $\wt{z}$ is temporarily omitted for simplicity, and is added back when necessary.} which maps from $(-\infty, \infty)$ to some target range building upon BP, 
\begin{align}\label{eqBPs}
h(\wt{z})&=\Psi^{-1}[B(\Phi(\wt{z}); k, \bd{\omega})], \cr B(u; k, \bd{\omega})&=\sum_{r=1}^{k}\omega_{r,k}I_{u}(r, k-r+1),
\end{align}
where $I_{u}(r, k-r+1)$ is the  regularized incomplete beta function.  $\Phi(\cdot)$ is the standard normal CDF mapping from $(-\infty, \infty)$ to $[0,1]$, and $\Psi^{-1}(\cdot)$ is some predefined tractable inverse CDF with fixed parameters; for example, the inverse CDF of the exponential distribution helps map from $[0,1]$ to $(0,\infty)$ for positive variables. $B(u; k, \bd{\omega})$ relocates the probability mass on the unit interval $[0,1]$.   The degree $k$ is an unknown smoothing parameter, and $\bd{\omega}$ is the unknown mixture weights on the probability simplex $\Delta_{k}=\{(\omega_{1}, \hdots, \omega_{k}): \omega_{i}\geq 0, \sum_{i}\omega_{i}=1\}$.  The proposed sandwich-type transformation avoids the difficulty of specifying any particular types of marginals,  while still leads to tractable derivations presented in Section \ref{sec:5}.


\subsection{Variational Inverse Transform}\label{secVIT}
 Considering a $\textrm{1-}$d variational approximation problem  ($x$ is a scalar, the true posterior $f(x)$  is known up to the normalizing constant), fix $q(\wt{z})=\cN(0,1)$, thus $u=\Phi(\wt{z})\sim\mathcal{U}[0,1]$,  we can learn the monotonic transformation $\xi(\cdot)=Q^{-1}(\cdot)$ on the base uniform distribution $q_{0}(u)$ by solving  a variational  problem,
\begin{align}
\xi^{\star}(\cdot) = \argmin_{\xi} \mathrm{KL}\{q(x)||f(x)\}, \quad  x=\xi(u)=Q^{-1}(u), \nonumber
\end{align}     
i.e.,   if we generate  $u \sim \mathcal{U}[0,1]$, then  $x=\xi^{\star}(u)\sim Q^{\star}$. $Q^{\star}$ is closest to the true distribution  $F$ with the minimum KL divergence. This can be interpreted as the variational counterpart of the inverse transform sampling \citep{Devroye86}, termed as variational inverse transform (VIT). Our BP-based construction  $\xi(\cdot)=Q^{-1}(\cdot)=\Psi^{-1}(B(u; k, \bd{\omega}))$ is one appropriate parameterization scheme for the inverse probability transformation $Q^{-1}(\cdot)$. VIT-BP offers two clear advantages.  First, as opposed to fixed-form variational Bayes, it does not require any specification of parametric form for $q(x)$. Second, the difficult task of calculating the general inverse CDFs $Q^{-1}(\cdot)$ is lessened to the much easier task of calculating the predefined tractable inverse  CDF $\Psi^{-1}(\cdot)$. Some choices of $\Psi(\cdot)$ include CDF of  $\cN(0,1)$ for variables in $(-\infty, \infty)$, $\mathrm{Beta}(2,2)$ for truncated variables in $(0,1)$.

%
%
%
To be consistent with  VIT, we shall set $\Phi(\cdot)$ in \eqref{eqBPs} to be $\Phi(\cdot|\mu, \sigma^{2})$,  instead of $\Phi(\cdot|0,1)$, such that $u$ is always uniformly distributed. Ideally, BP itself suffices to represent arbitrary continuous distribution function on the unit interval.  However, it might require a higher order $k$.  As is demonstrated in Section \ref{sectionbp}, this requirement can be alleviated by incorporating auxiliary parameters $\{\mu, \sigma^{2}\}$ in VGC-BP, which potentially help in changing location and dispersion of the probability mass. 

\section{Stochastic VGC}\label{sec:5}

\begin{algorithm*}[t]
   \caption{(VGC-BP) Stochastic Variational Gaussian Copula Inference with Bernstein Polynomials}
   \label{alg:example}
\begin{algorithmic}
   \STATE {\bfseries Input:} observed data $\bd{y}$, user specified model $\ln p(\bd{y},\bd{x})$ and first-order derivatives $\nabla_{\bd{x}}\ln p(\bd{y}, \bd{x})$, Bernstein polynomials degree $k$, predefined $\Psi(\cdot)$ and $\Phi(\cdot)$  \STATE \textbf{Initialize} variational parameter $\Theta_{0}=\left(\bd{\mu}_{0}, \bd{C}_{0}, \{\bd{\omega}^{(j)}_{0}\}_{j=1:p}\right)$, $t=0$.
   \REPEAT
   \STATE  $t=t+1$,
   \STATE  Sample $\wt{\bd{\epsilon}}\sim q_{\mathrm{G}}\left(\bd{\wt{\epsilon}}, \bd{0}, \bd{I}_{p}\right)$,  and set $\wt{\bd{z}}=\bd{\mu}_{t-1}+\bd{C}_{t-1}\bd{\epsilon}$,
   \STATE  $\bd{\mu}_{t} = \bd{\mu}_{t-1}+\lambda_{t}[\nabla_{\wt{{\bd{z}}}}\ell_{s}(\wt{\bd{z}}, {h})-\nabla_{\wt{\bd{z}}}\ln{q_{\mathrm{G}}(\wt{\bd{z}})}]$,  \hfill \% Update $\bd{\mu}_{t-1}$ with stepsize  $\lambda_{t}$ 
   \STATE $\bd{C}_{t} = \bd{C}_{t-1}+\eta_{t}[\nabla_{\wt{{\bd{z}}}}\ell_{s}(\wt{\bd{z}}, {h})-\nabla_{\wt{\bd{z}}}\ln{q_{\mathrm{G}}(\wt{\bd{z}})}]\bd{\epsilon}^{T}
$, \hfill \% Update $\bd{C}_{t-1}$ with stepsize  $\eta_{t}$ 
   \FOR{$j=1$ {\bfseries to} $p$}
\STATE $\bd{\omega}^{(j)}_{t} = \mathcal{P}(\bd{\omega}^{(j)}_{t-1} +\xi_{t}^{(j)}\nabla_{\bd{\omega}^{(j)}}\ell_{s}(\wt{\bd{z}}, h))$,  \hfill \% Update $\bd{\omega}^{(j)}_{t-1}$ with stepsize  $\xi_{t}^{(j)}$ and gradient projection $\mathcal{P}$ 
   \ENDFOR
   \UNTIL{convergence criterion is satisfied}
   \STATE {\bfseries Output:}  marginal parameters $\left(\{\bd{\omega}^{(j)}\}_{j=1:p},\bd{\mu},\bd{\sigma^{2}}\right)$ and copula parameters $\bd{\Upsilon}$
\end{algorithmic}
\end{algorithm*}

The derivations of deterministic VGC updates are highly model-dependent. First, due to the cross terms often involved in the log likelihood/prior,  the corresponding Gaussian expectations and their derivatives may not be analytically tractable.  Second, owing to the non-convex nature of many problems,  only locally optimal solutions can be guaranteed. In contrast, stochastic implementation of VGC only requires the evaluation of the log-likelihood and log-prior along with their derivatives, eliminating most model-specific derivations, and it provides a chance of escaping local optima by introducing randomness in gradients.
 

\subsection{Coordinate transformations}\label{secmuC}
Applying the coordinate transformations\footnote{If necessary, the Gaussian copula can be replaced with other appropriate parametric forms. The coordinate transformation supports many other distributions as well, for example, those described in Appendix C.2. of Rezende et al. (2014).
} of stochastic updates, $\wt{\bd{z}}=\bd{\mu}+\bd{C}\bd{\epsilon}$, $\bd{\epsilon}\sim \cN(\bd{0}, \bd{I})$,  introduced in  \citep{rezende2014stochastic, titsias2014doubly}, the gradient of the ELBO w.r.t. variational parameter $(\bd{\mu}, \bd{C})$ can be written as  
\begin{align}\label{ourscheme}
\hspace{-0.65cm}\nabla_{\bd{\mu}}\mathcal{L}_{\mathrm{C}}&=\mathbb{E}_{q_{G} (\wt{\bd{z}})}\left[\nabla_{\wt{\bd{z}}}\ell_{s}(\wt{\bd{z}}, h)-\nabla_{\wt{\bd{z}}}\ln{q_{\mathrm{G}}(\wt{\bd{z}})}\right],\cr \hspace{-0.6cm}
\nabla_{\bd{C}}\mathcal{L}_{\mathrm{C}}&=\mathbb{E}_{q_{G}(\wt{\bd{z}})}\left[\nabla_{\wt{\bd{z}}}(\ell_{s}(\wt{\bd{z}}, h)-\nabla_{\wt{\bd{z}}}\ln{q_{\mathrm{G}}(\wt{\bd{z}})})\bd{\epsilon}^{T}\right],
\end{align}
where the stochastic gradient terms 
\begin{align}
\nabla_{\wt{{z}_{j}}}\ell_{s}(\wt{\bd{z}})&=\nabla_{\wt{{z}_{j}}}  \ln p(\bd{y},h(\wt{\bd{z}}))+\nabla_{\wt{{z}_{j}}}\ln h_{j}'(\wt{z}_{j})\cr &={\color{black}\frac{\partial \ln p(\bd{y},\bd{x})}{\partial x_{j}}}h_{j}'({\wt{{z}_{j}}})+\nabla_{\wt{{z}_{j}}}\ln h_{j}'(\wt{z}_{j}). \nonumber
\end{align}

According to the chain rule, the first derivative of $h(\cdot)$ w.r.t $\wt{z}$ is, 
\begin{align}\label{eq8}
h'(\wt{z})&=\frac{d \Psi^{-1}[B(\Phi(\wt{z}); k, \bd{\omega})]}{d B(\Phi(\wt{z}); k, \bd{\omega})}\frac{d B(\Phi(\wt{z}); k, \bd{\omega})}{d \Phi(\wt{z})}\frac{d \Phi(\wt{z})}{d \wt{z}}\nonumber \\& = \frac{b(\Phi(\wt{z}); k, \bd{\omega})\phi(\wt{z})}{\psi(h(\wt{z}))},
\end{align}
where  $b(u; k, \bd{\omega})=\sum_{r=1}^{k}\omega_{r,k}\beta(u; r, k-r+1)$,  $\beta(x; a, b)$ is the beta density $\beta(x; a, b)={\Gamma(a+b)}/{(\Gamma(a)\Gamma(b))}x^{a-1}(1-x)^{b-1}$. Therefore,  $\ln h'(\wt{z})= \ln b(\Phi(\wt{z}); k, \bd{\omega})+\ln \phi(\wt{z}) -\ln\psi(h(\wt{z}))$  and $\nabla_{\wt{{z}_{j}}}\ln h_{j}'(\wt{z}_{j})={h_{j}''(\wt{z}_{j})}/{h_{j}'(\wt{z}_{j})} $ all take analytical expressions, where 
\begin{align}
h_{j}''(\wt{z}_{j})&=[\rho_{1}'(\wt{z}_{j})\rho_{2}(\wt{z}_{j})\rho_{3}(\wt{z}_{j})+\rho_{1}(\wt{z}_{j})\rho_{2}'(\wt{z}_{j})\rho_{3}(\wt{z}_{j})\cr &-\rho_{1}(\wt{z}_{j})\rho_{2}(\wt{z}_{j})\rho_{3}'(\wt{z}_{j})]/{[\rho_{3}(\wt{z}_{j})]^{2}},\nonumber
\end{align}
where $\rho_{1}(\wt{z}_{j})=b(u_{j}; k, \bd{\omega}^{(j)})$, $\rho_{2}(\wt{z}_{j})=\phi(\wt{z_{j}})$, $\rho_{3}(\wt{z}_{j})=\psi(h_{j}(\wt{z}_{j}))$, $\rho_{1}'(\wt{z}_{j})=\phi(\wt{z_{j}})\sum_{r=1}^{k}\omega_{r,k}^{(j)}\beta'(u_{j}; r, k-r+1)$,  $\rho_{2}'(\wt{z}_{j})=-\wt{z_{j}}\phi(\wt{z_{j}})$,  $\rho_{3}'(\wt{z}_{j})=\psi'(h_{j}(\wt{z}_{j}))h_{j}'(\wt{z}_{j})$, $u_{j}=\Phi(\wt{z_{j}})$, $\phi(\cdot)$ is the PDF of $\cN(0,1)$, $\psi(\cdot)$ and $\psi'(\cdot)$ are the predefined PDF and its derivative respectively. 
 Defining $\beta(x; a, 0)=\beta(x; 0, b)=0$, the derivative is written as a combination of two polynomials of lower degree
$\beta'(x; a, b)=(a+b-1)[\beta(x; a-1, b)-\beta(x; a, b-1)]$. 

In stochastic optimization, the gradients expressed in terms of expectations are approximated using Monte Carlo integration with finite samples. The gradients
contain expectations on additive terms. Note that  \cite{rezende2014stochastic} and \cite{titsias2014doubly} ignore the stochasticity in the entropy term $\mathbb{E}_{q_{G} (\wt{\bd{z}})}[-\ln{q_{\mathrm{G}}(\wt{\bd{z}})]}$ and  assume $\nabla_{\bd{\mu}}\mathbb{E}_{q_{G} (\wt{\bd{z}})}[-\ln{q_{\mathrm{G}}(\wt{\bd{z}})]}=0$ and  $\nabla_{\bd{C}}\mathbb{E}_{q_{G} (\wt{\bd{z}})}[-\ln{q_{\mathrm{G}}(\wt{\bd{z}})]}=\diag[1/C_{jj}]_{j=1:p}$.  This creates an inconsistency as we only take finite samples in approximating $\mathbb{E}_{q_{G} (\wt{\bd{z}})}[\nabla_{\wt{\bd{z}}}\ell_{s}(\wt{\bd{z}})]$, and perhaps surprisingly, this also results in an increase of  the gradient variance  and the sensitivity to the learning rates. Our method is inherently more stable, as the difference between the gradients, $\nabla_{\wt{\bd{z}}}[\ell_{s}(h(\wt{\bd{z}}))-q_{\mathrm{G}}(\wt{\bd{z}})]$, $\forall \wt{\bd{z}}$, tends to zero when the convergent point is approached. In contrast, the gradients in previous method diffuses with a constant variance even around the global maximum. This phenomenon is illustrated in Section \ref{sec2dLN}. 

The alternative log derivative approach are also applicable to VGC inference and other types of copulas, see \cite{PaisleyBlei, mnih2014neural, rezende2014stochastic} for references. We leave this exploration open for future investigation.

\subsection{Update the BP Weights}

Under a given computational budget, we prefer a higher degree $k$, as there is no over-fitting issue in this variational density approximation task.   
 Given $k$, the basis functions are completely known, depending only on index $r$. The only parameter left to be optimized in the Bernstein polynomials is the mixture weights. Therefore, this construction is relatively simpler than Gaussian mixture proposals \citep{GershmanBlei, nguyen2014automated}. Assuming permissibility of  
interchange of integration and differentiation holds, we have  $\nabla_{\bd{\omega}^{(j)}}\mathcal{L}_{\mathrm{C}}=\mathbb{E}_{q_{G}(\wt{\bd{z}})}\left[\nabla_{\bd{\omega}^{(j)}}\ell_{s}(\wt{\bd{z}}, h, \bd{y})\right]$, with the stochastic gradients
\begin{align}
& \nabla_{\bd{\omega}^{(j)}}\ell_{s}(\wt{\bd{z}}, h, \bd{y})= \nabla_{\bd{\omega}^{(j)}}\ln p(\bd{y},h(\wt{\bd{z}}))+\nabla_{\bd{\omega}^{(j)}} \ln h_{j}'(\wt{z_{j}})
\cr &=\frac{\partial \ln p(\bd{y},\bd{x})}{\partial x_{j}}\bigg[\frac{\partial h_{j}(\wt{z}_{j})}{\partial \omega_{r,k}^{(j)}}\bigg]_{r=1:k} + \bigg[\frac{\partial \ln h_{j}'(\wt{z_{j}})}{\partial\omega_{r,k}^{(j)}}\bigg]_{r=1:k}, \nonumber
\end{align}  
where
\begin{align}
\frac{\partial h_{j}(\wt{z}_{j})}{\partial \omega_{r,k}^{(j)}}&=\frac{\partial \Psi^{-1}[B(u_{j}; k, \bd{\omega}^{(j)})]}{\partial \omega_{r,k}^{(j)}} 
=\frac{I_{u_{j}}(r, k-r+1)}{\psi( h_{j}(\wt{z}_{j}))},\nonumber
\end{align}
\begin{align}
{\partial \ln h_{j}'(\wt{z_{j}})}/{\partial\omega_{r,k}^{(j)}}&={\beta(u_{j}; r, k-r+1)}/{b(u_{j}; k, \bd{\omega}^{(j)})}\cr &-\frac{\psi'(h_{j}(\wt{z}_{j}))}{\{\psi(h_{j}(\wt{z}_{j}))\}^{2}}I_{u_{j}}(r, k-r+1). \nonumber
\end{align}
The gradients w.r.t $\bd{\omega}^{(j)}$ turn into expectation straightforwardly,  to enable stochastic optimization of the ELBO.  To satisfy the constraints of $\bd{\omega}^{(j)}$ on the probability simplex, we apply the gradient projection operation $\mathcal{P}$  introduced in \cite{duchi2008efficient} with complexity $\mathcal{O}(k\log{k})$.  The above derivations related to BPs  together with those in Section \ref{secmuC} are all analytic and model-independent. The only two  model-specific terms are  ${\color{black}\ln p(\bd{y},\bd{x})}$ and  ${\color{black}{\partial \ln p(\bd{y},\bd{x})}/{\partial \bd{x}}}$.  The stochastic optimization algorithm is  summarized in Algorithm \ref{alg:example}, with little computational overhead  added relative to stochastic VG.  The stability and efficiency of the stochastic optimization algorithm can be further improved by embedding adaptive subroutines   \citep{duchi2011adaptive} and considering second-order optimization method \citep{fankai2015}.

\section{Experiments}
 We use Gaussian copulas with fixed/free-form margins as automated \emph{inference engines} for posterior approximation in generic hierarchical Bayesian models. We evaluate  the peculiarities reproduced in the univariate margins and the posterior dependence captured broadly across latent variables. This is done by comparing VGC methods to the ground truth and other baseline methods such as MCMC, MFVB, and VG (see Supplementary Material for detailed derivations). Matlab code for VGC is available from the GitHub repository: \url{https://github.com/shaobohan/VariationalGaussianCopula}
 
\vspace{-0.08cm}
 \subsection{Flexible Margins}\label{sectionbp}

\begin{figure}[bpht] 
\vskip -0.05in
\begin{center}
\tiny{
$\begin{array}{cc}
\hspace{-0.5cm}\includegraphics[height=0.16\textwidth, width=0.2\textwidth]{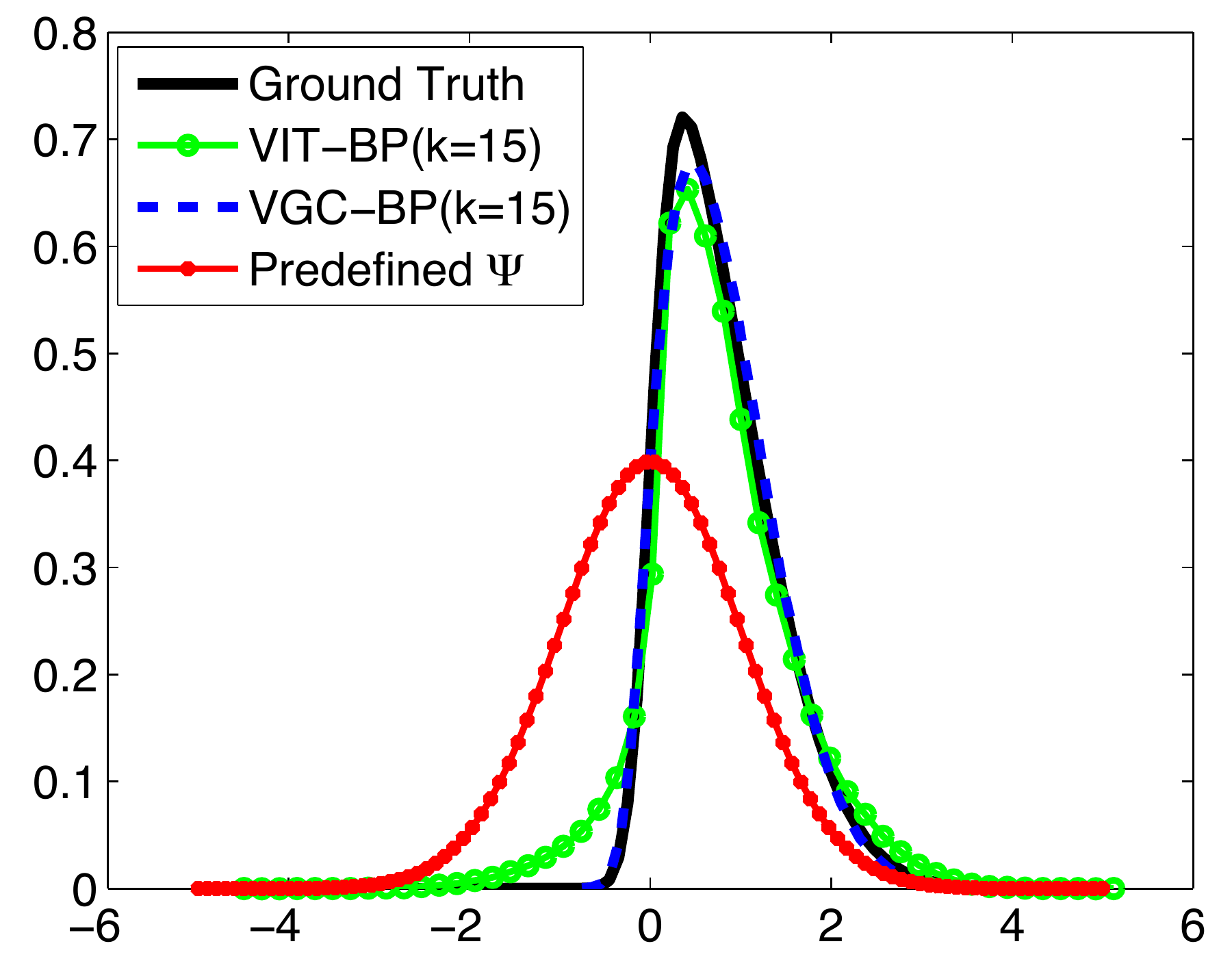} &
\hspace{-0.4cm}\includegraphics[height=0.16\textwidth, width=0.2\textwidth]{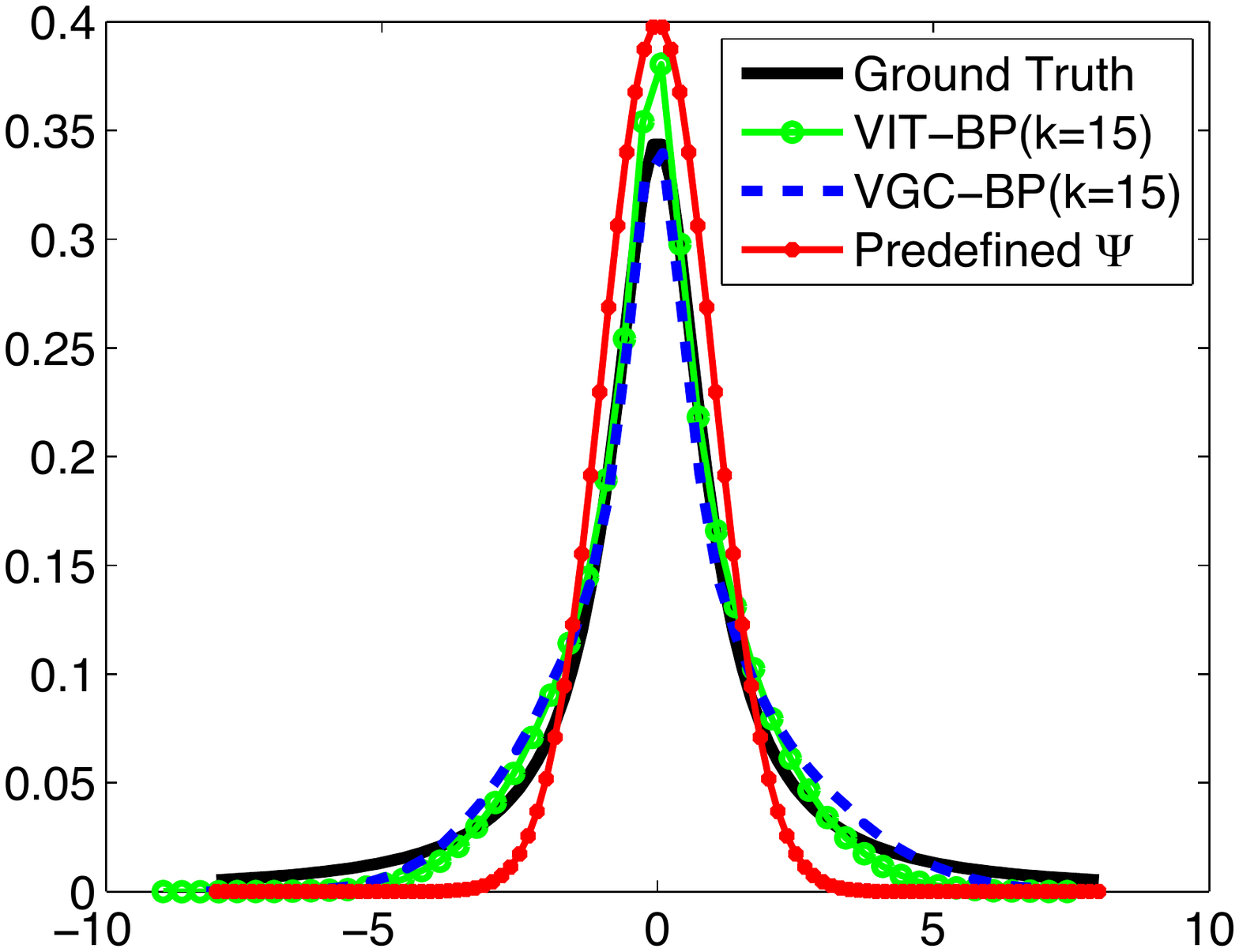}\\
(a) \textrm{ Skew-Normal} (\alpha = 5)& (b)\textrm{ Student's t } (\nu = 1)\\
\hspace{-0.5cm}\includegraphics[height=0.16\textwidth, width=0.2\textwidth]{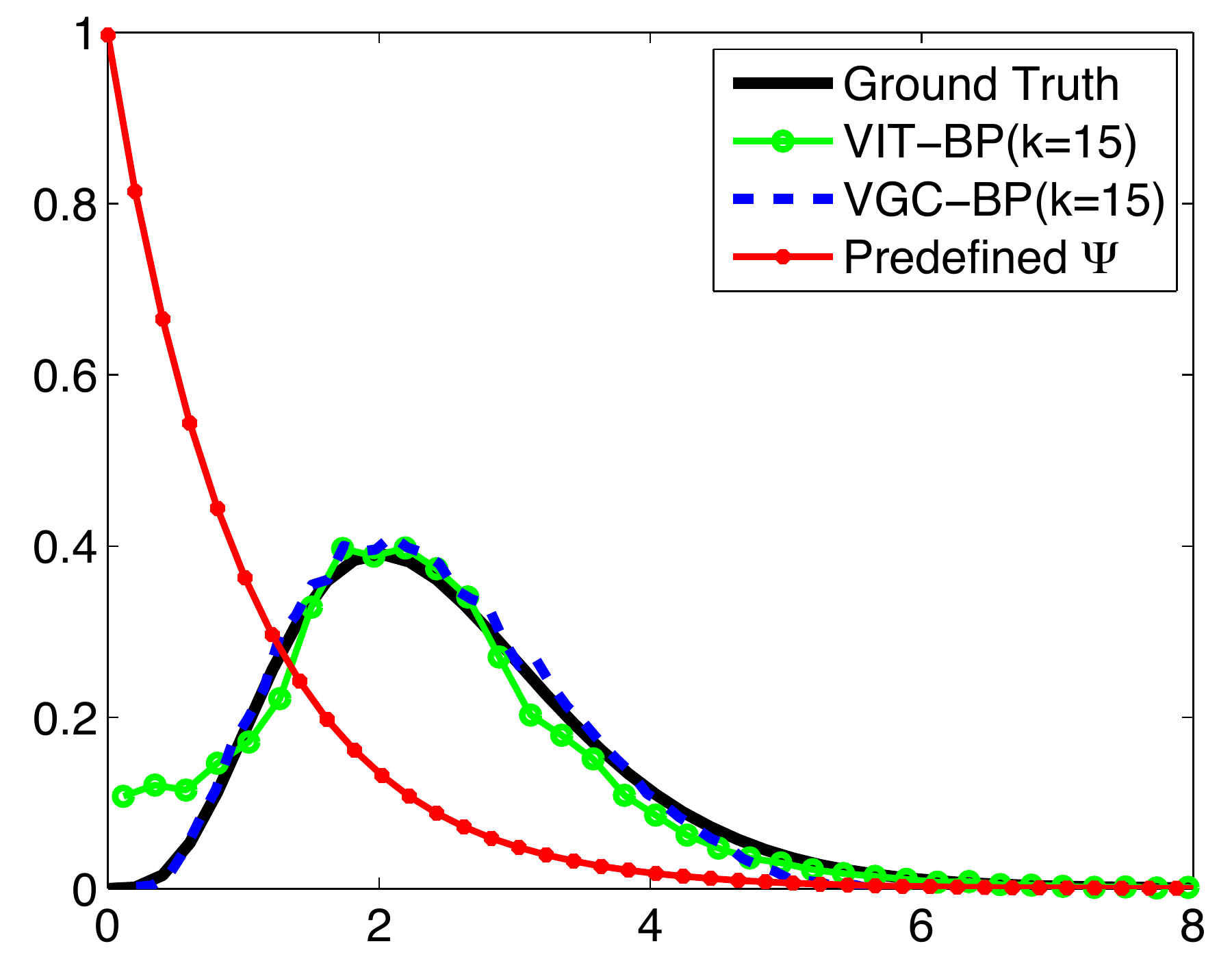} &
\hspace{-0.4cm}\includegraphics[height=0.16\textwidth, width=0.2\textwidth]{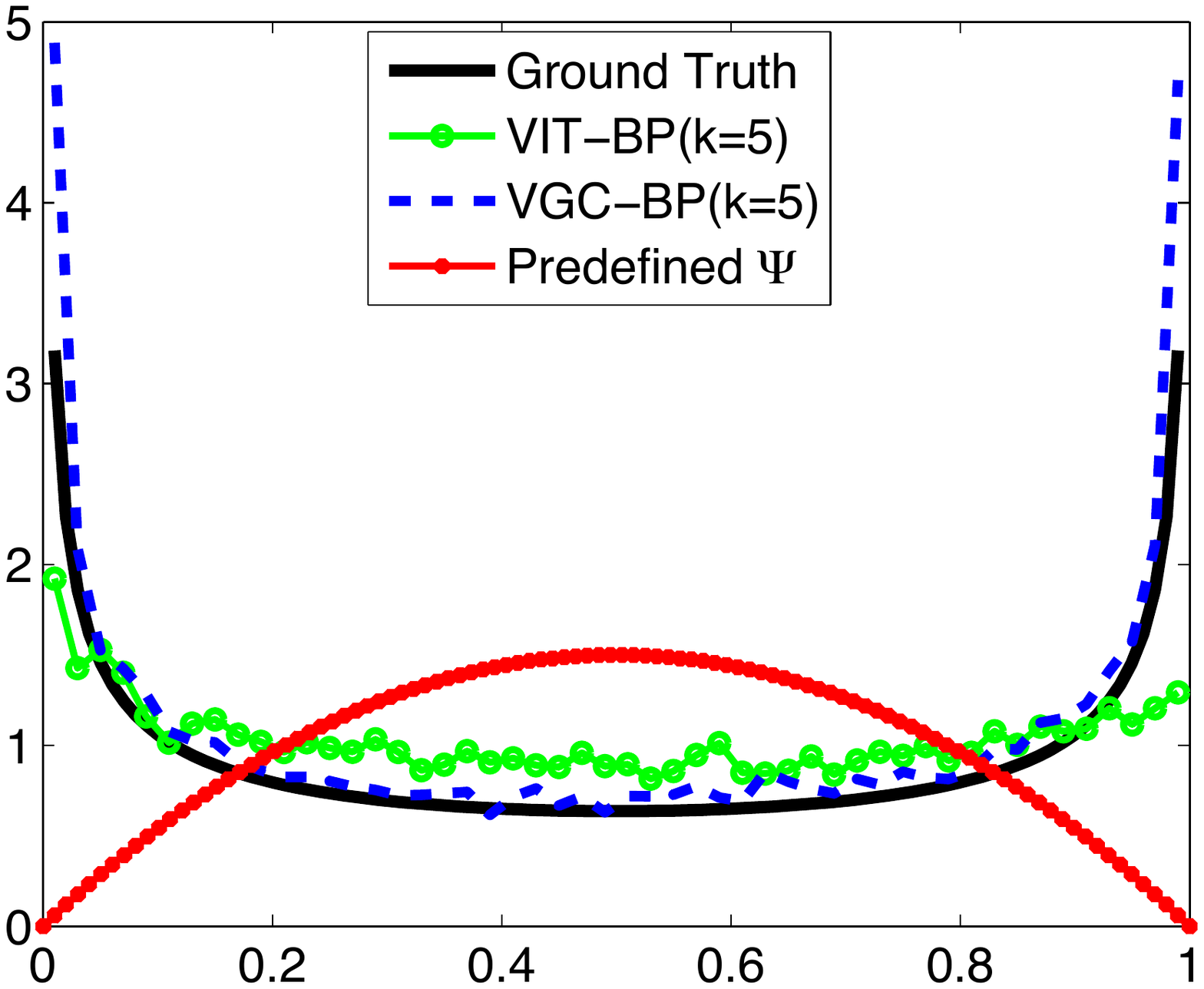}\\
(c) \textrm{ Gamma}(5,2)& (d) \textrm{ Beta}(0.5, 0.5)
\end{array}$}
\end{center}
\vskip -0.15in
\caption{Marginal Adaptation: VIT-BP v.s. VGC-BP}
\label{fig1d}
\end{figure}

We first assess the marginal approximation accuracy of our BP-based constructions in Section \ref{secVIT}, i.e., $h(\cdot)=\Psi^{-1}(B(\Phi(\wt{z}); k, \bd{\omega}))$ via $1\mhyphen$d variational optimization, where $\wt{z}\sim\cN(0,1)$ in VIT-BP, and $\wt{z}\sim\cN(\mu,\sigma^{2})$ in VGC-BP. For fixed BP order $k$, the shape of $q(x)$ is adjusted solely by updating $\bd{\omega}$, according to the variational rule. In VGC-BP, the additional marginal parameters $\{\mu, \sigma^{2}\}$ also contribute in changing location and dispersion of $q(x)$.  Examining Figure \ref{fig1d}, VGC-BP produces more accurate densities  than VIT-BP under the same order $k$.  Hereafter, the predefined $\Psi(\cdot)$ for real variables, positive real variable, and truncated [0,1] variables are  chosen to be the CDF of $\cN(0,1)$, $\mathrm{Exp}(1)$ and $\mathrm{Beta}(2,2)$, respectively. 

\subsection{Bivariate Log-Normal}\label{sec2dLN}
The bivariate log-normal PDF $p(x_{1}, x_{2})$ \citep{aitchison1957lognormal} is given by 
\begin{align}
p(x_{1}, x_{2})&={\exp{(-{\zeta}/{2})}}/[{2\pi x_{1} x_{2} \sigma_{{1}}\sigma_{{2}}\sqrt{1-\rho^2}}],\cr 
\zeta &= \frac{1}{1-\rho^2}\bigg[\alpha_{1}^{2}(x_{1})-2\rho\alpha_{1}(x_{1})\alpha_{2}(x_{2})+\alpha_{2}^{2}(x_{2})\bigg],\nonumber
\end{align} 
 where $\alpha_{i}(x_{i})={(\ln{x_{i}}-\mu_{i})}/{\sigma_{i}}$, $ i = 1, 2$, $-1<\rho<1$.   
  
\begin{figure}[bpht]
\vskip -0.07in
\begin{center}
\tiny{
$\begin{array}{ccccc} 
\hspace{-0.35cm}\includegraphics[height=0.09\textwidth, width=0.1\textwidth]{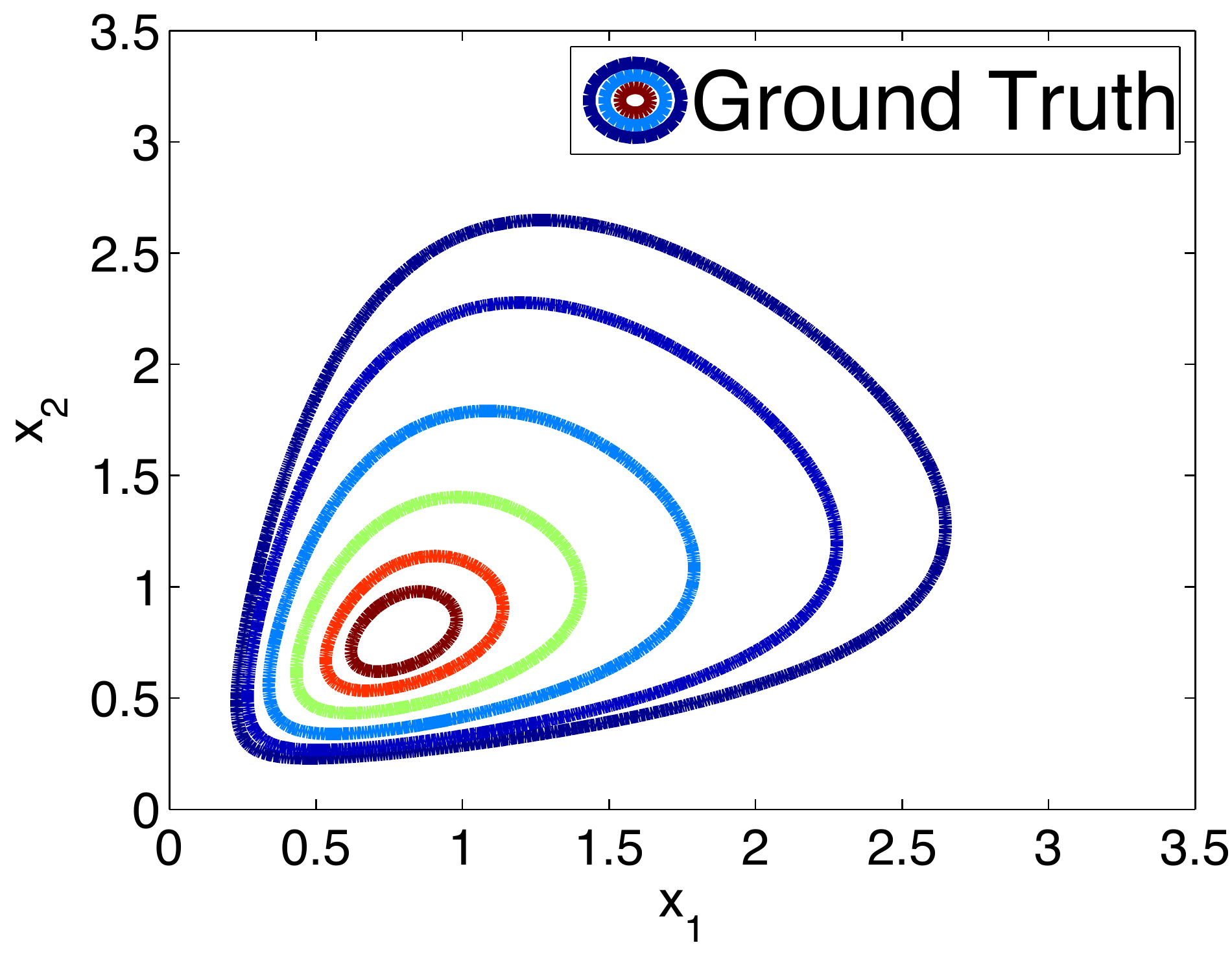} &
\hspace{-0.35cm}\includegraphics[height=0.09\textwidth, width=0.1\textwidth]{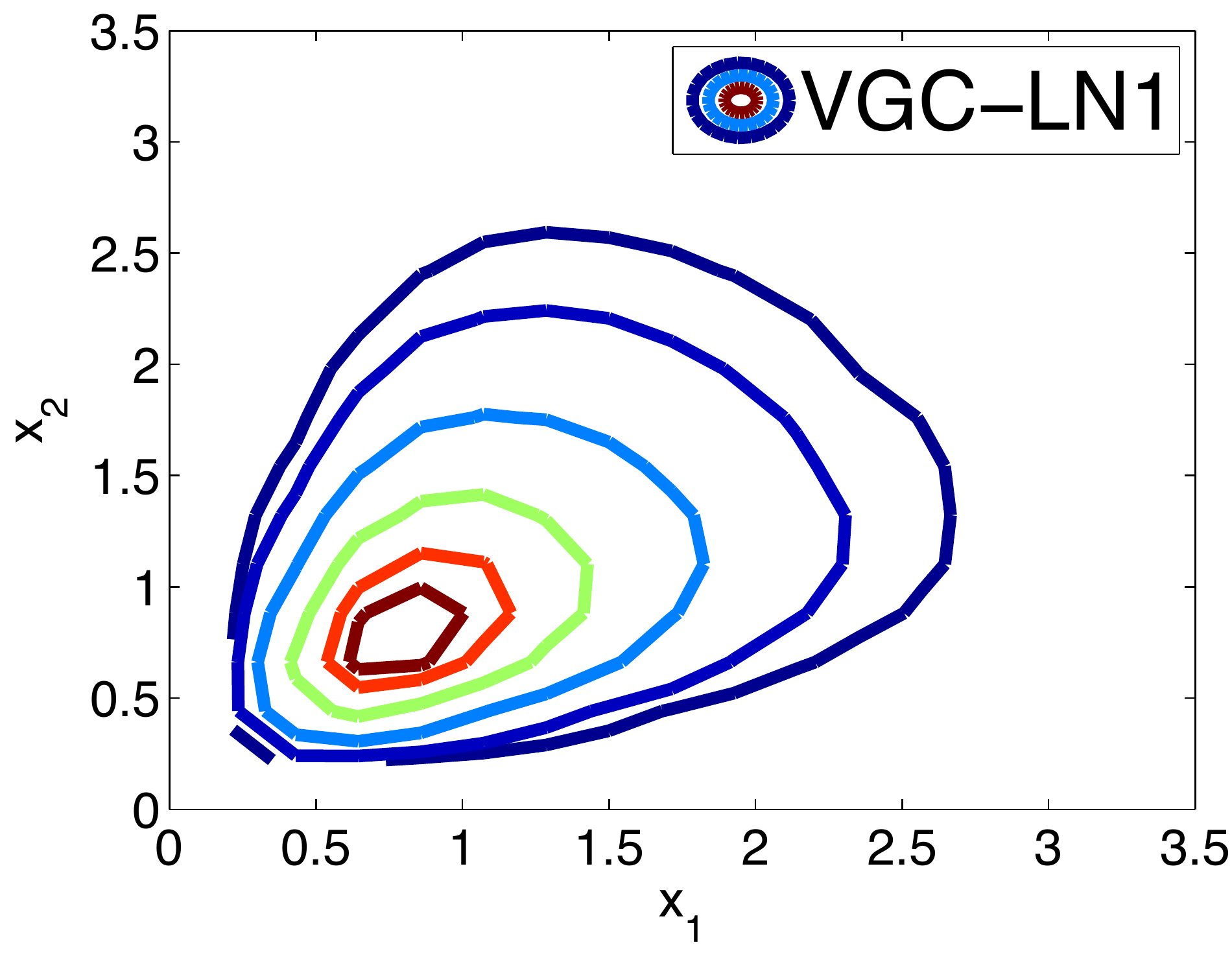} &
\hspace{-0.35cm}\includegraphics[height=0.09\textwidth, width=0.1\textwidth]{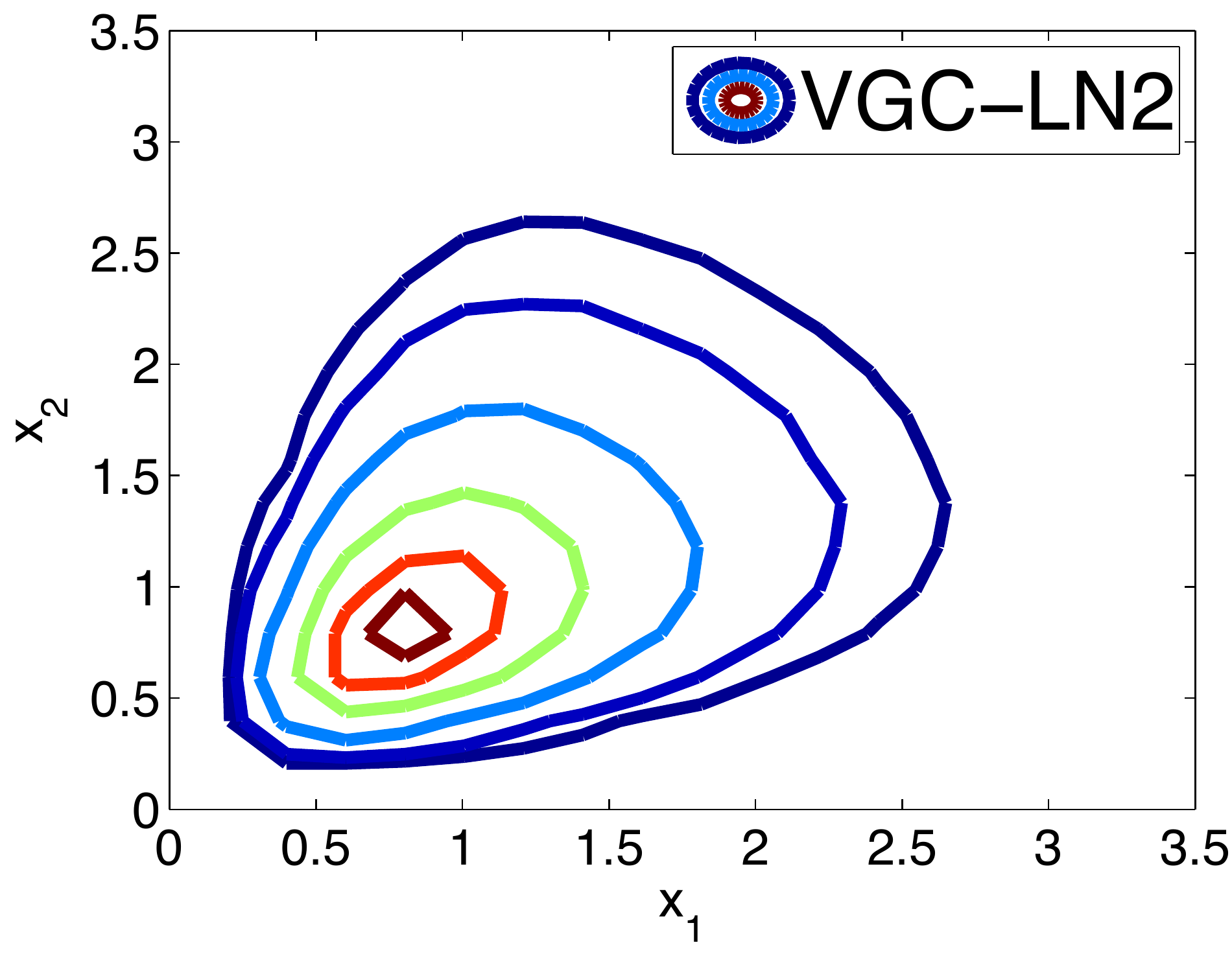} &
\hspace{-0.35cm}\includegraphics[height=0.09\textwidth, width=0.1\textwidth]{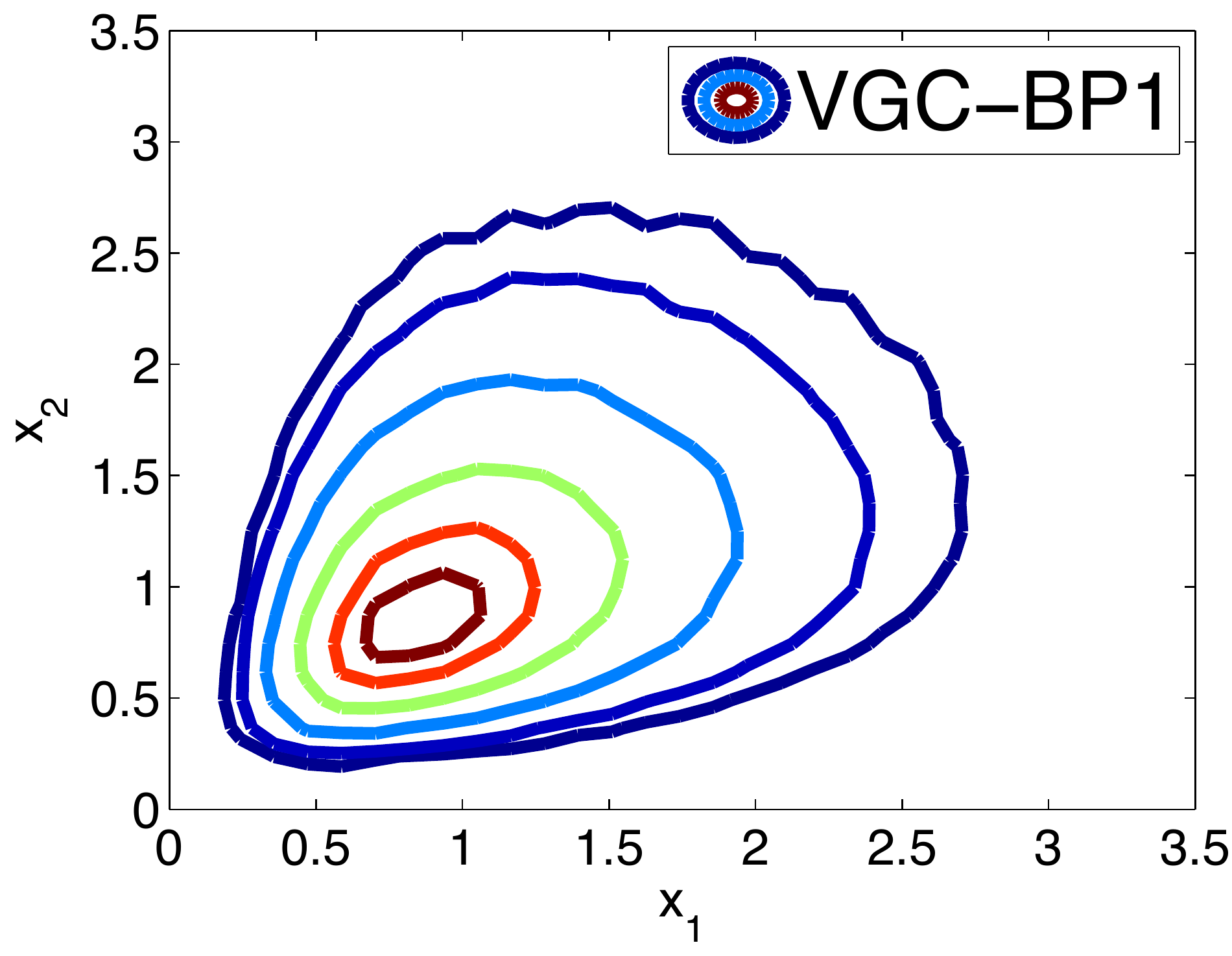} &
\hspace{-0.35cm}\includegraphics[height=0.09\textwidth, width=0.1\textwidth]{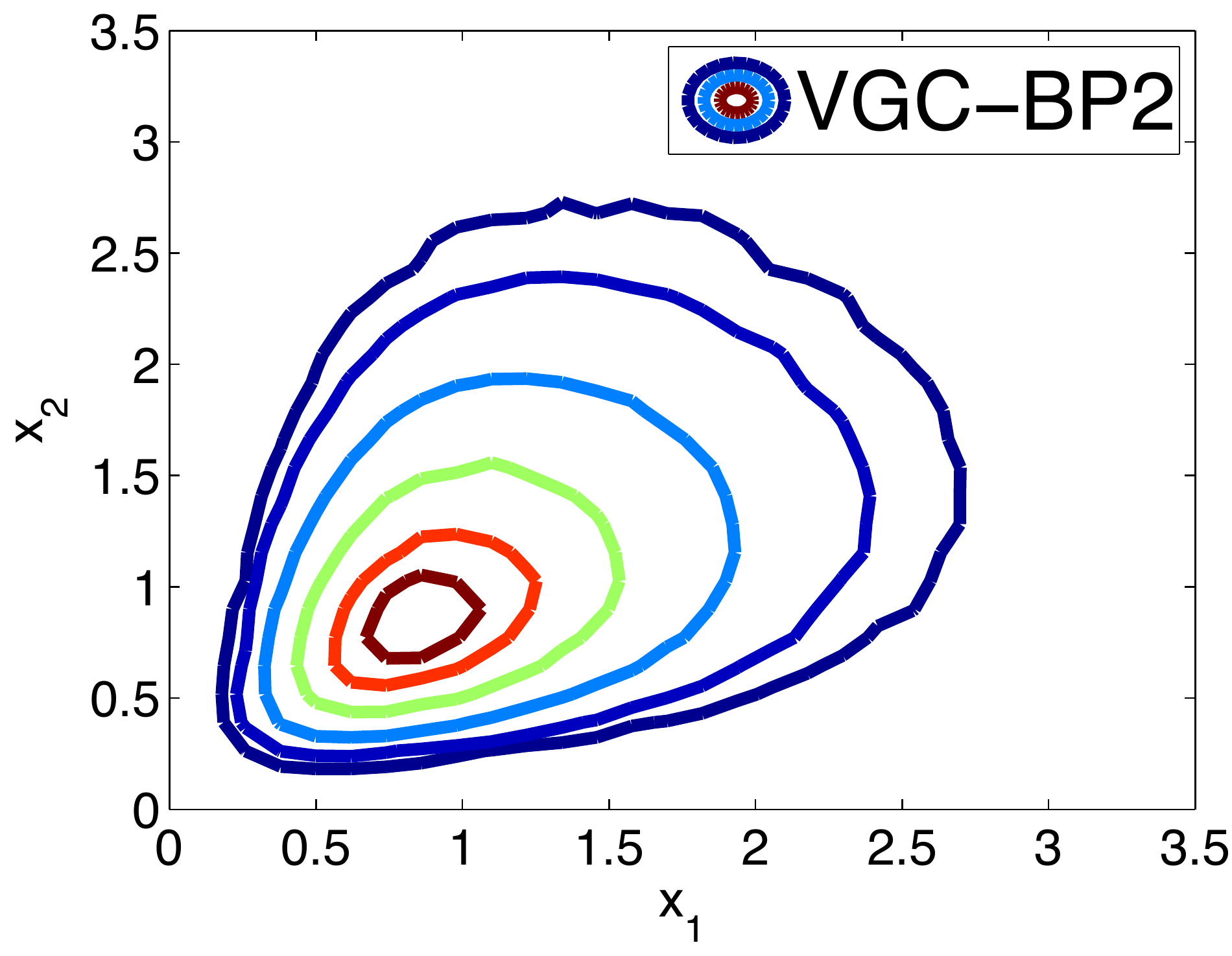}\\[-0.2em] 
\hspace{-0.35cm}\includegraphics[height=0.09\textwidth, width=0.1\textwidth]{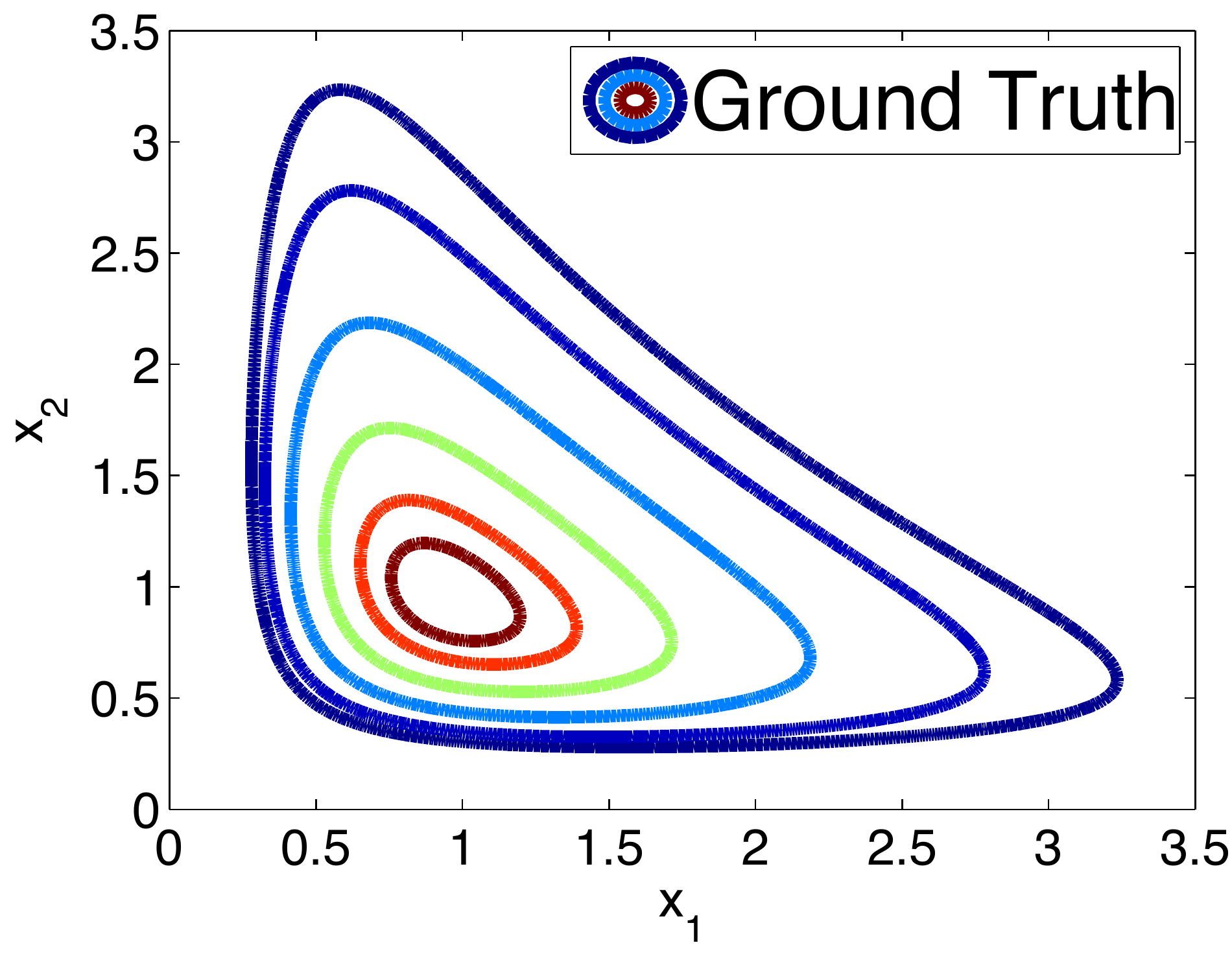} &
\hspace{-0.35cm}\includegraphics[height=0.09\textwidth, width=0.1\textwidth]{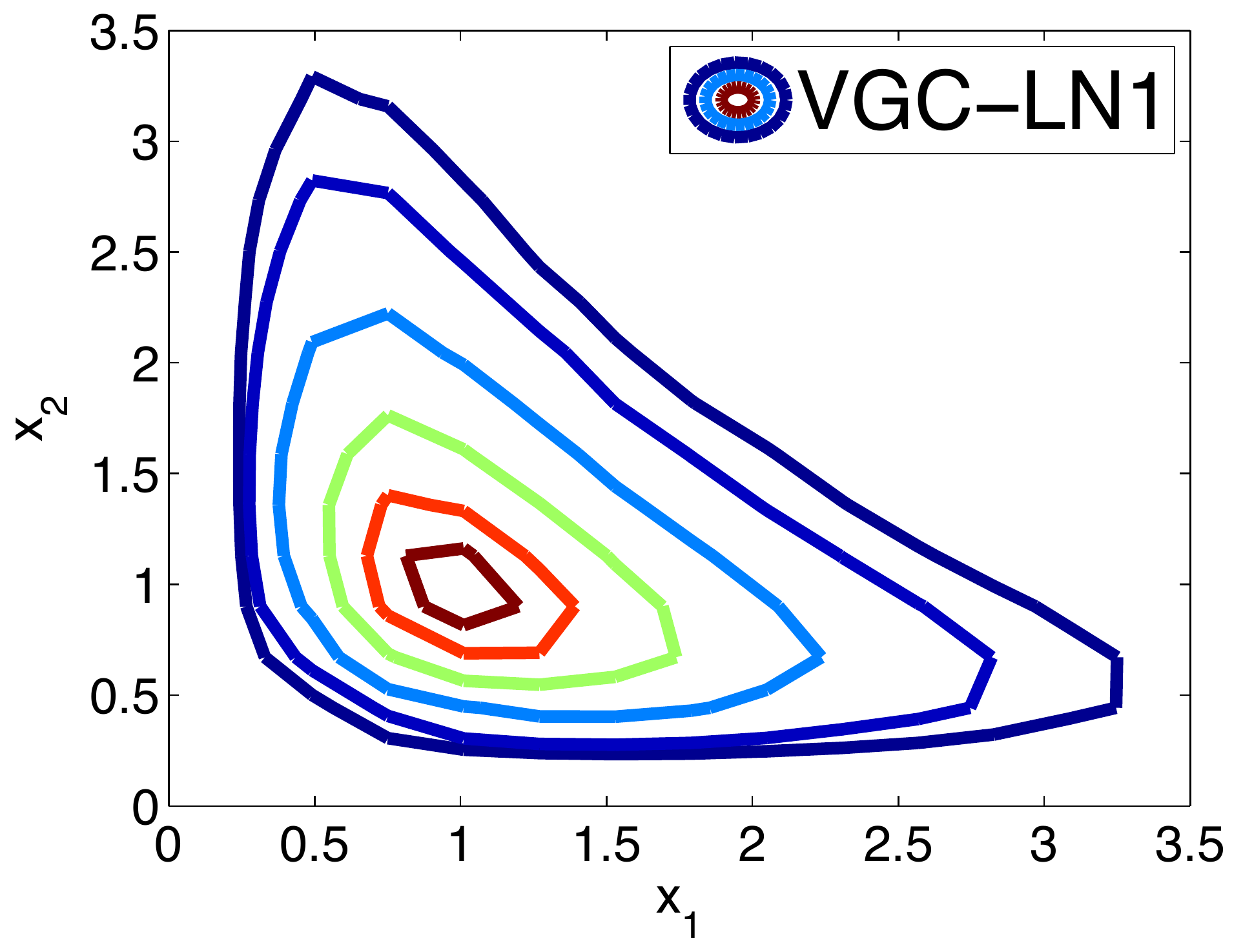} &
\hspace{-0.35cm}\includegraphics[height=0.09\textwidth, width=0.1\textwidth]{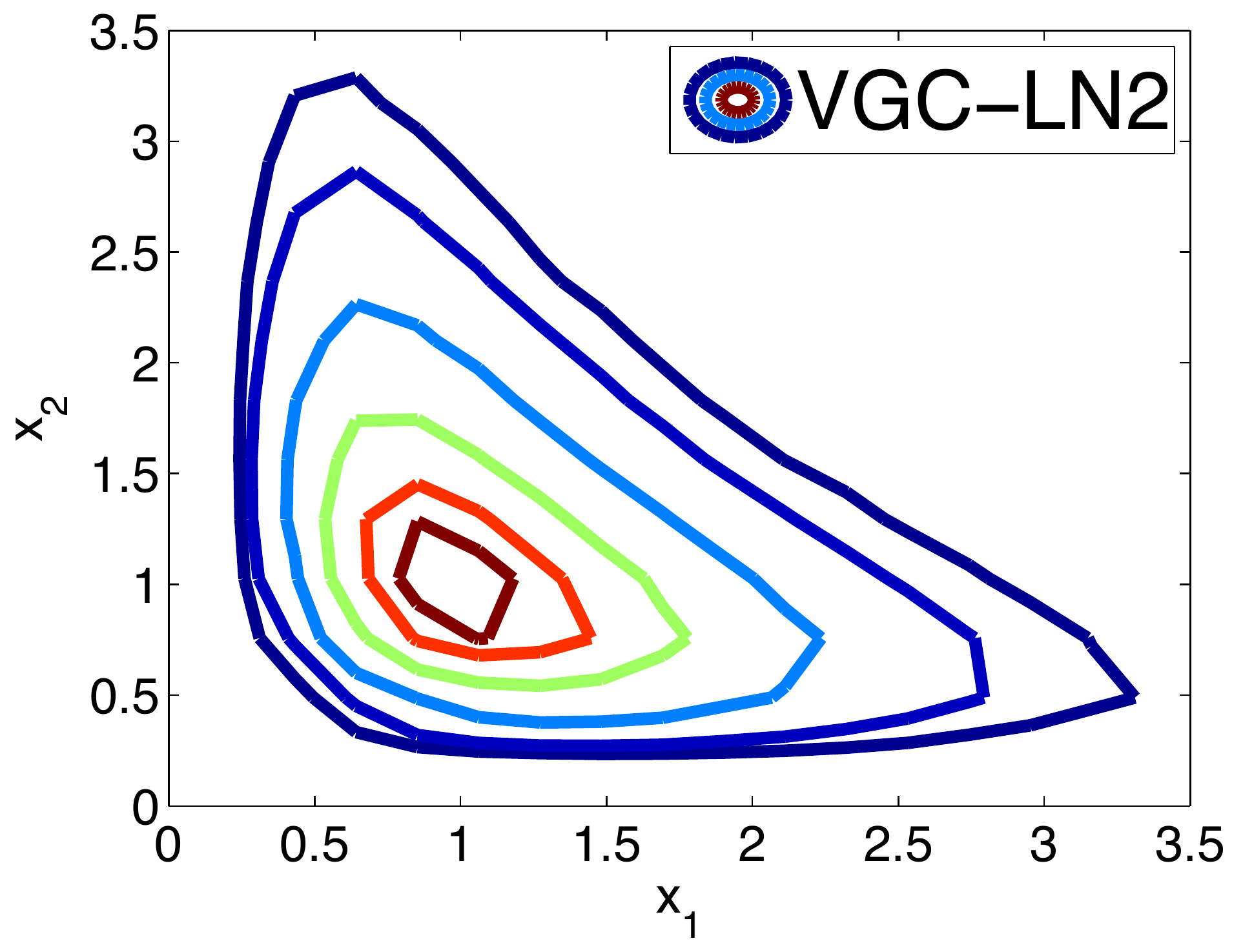} &
\hspace{-0.35cm}\includegraphics[height=0.09\textwidth, width=0.1\textwidth]{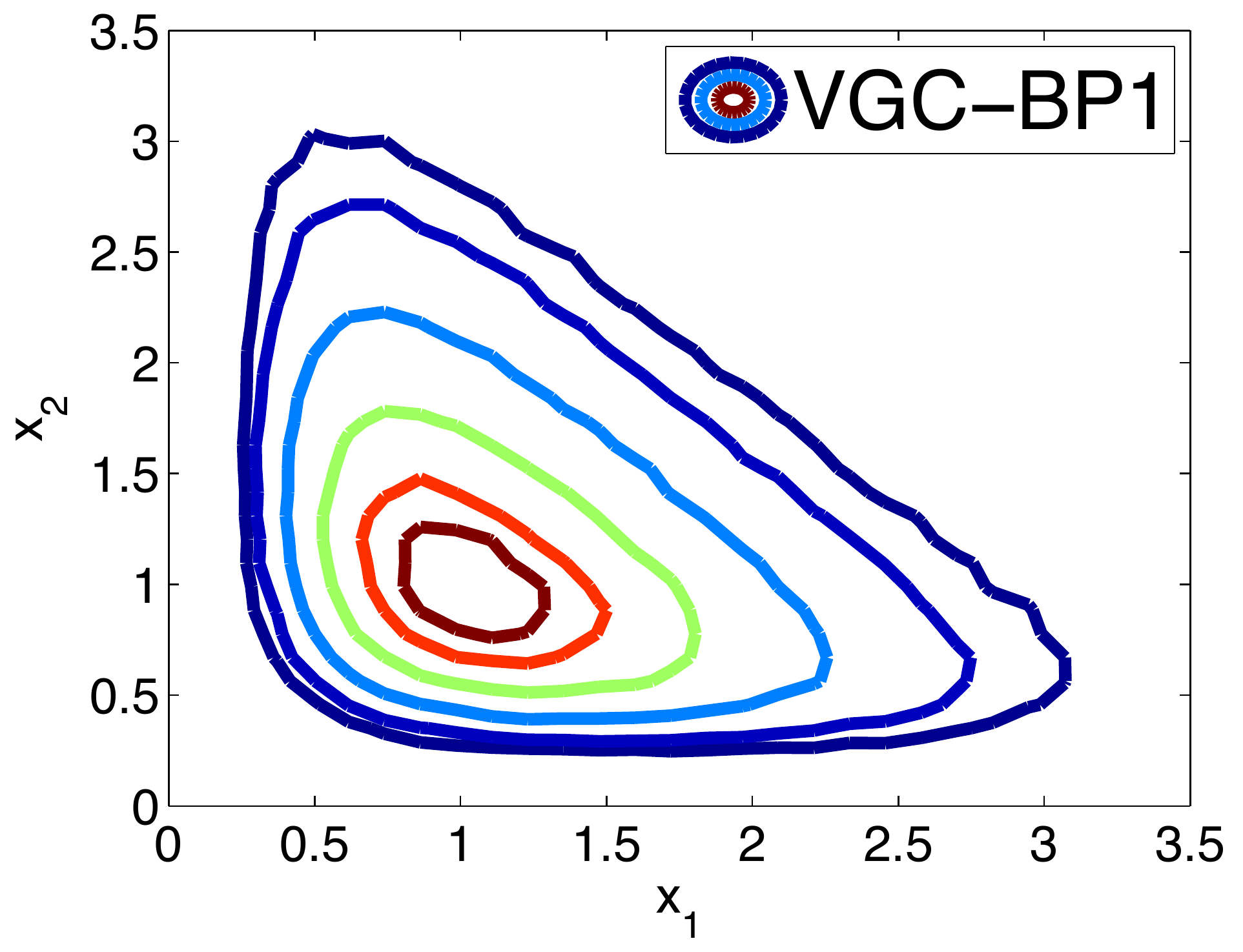} &
\hspace{-0.35cm}\includegraphics[height=0.09\textwidth, width=0.1\textwidth]{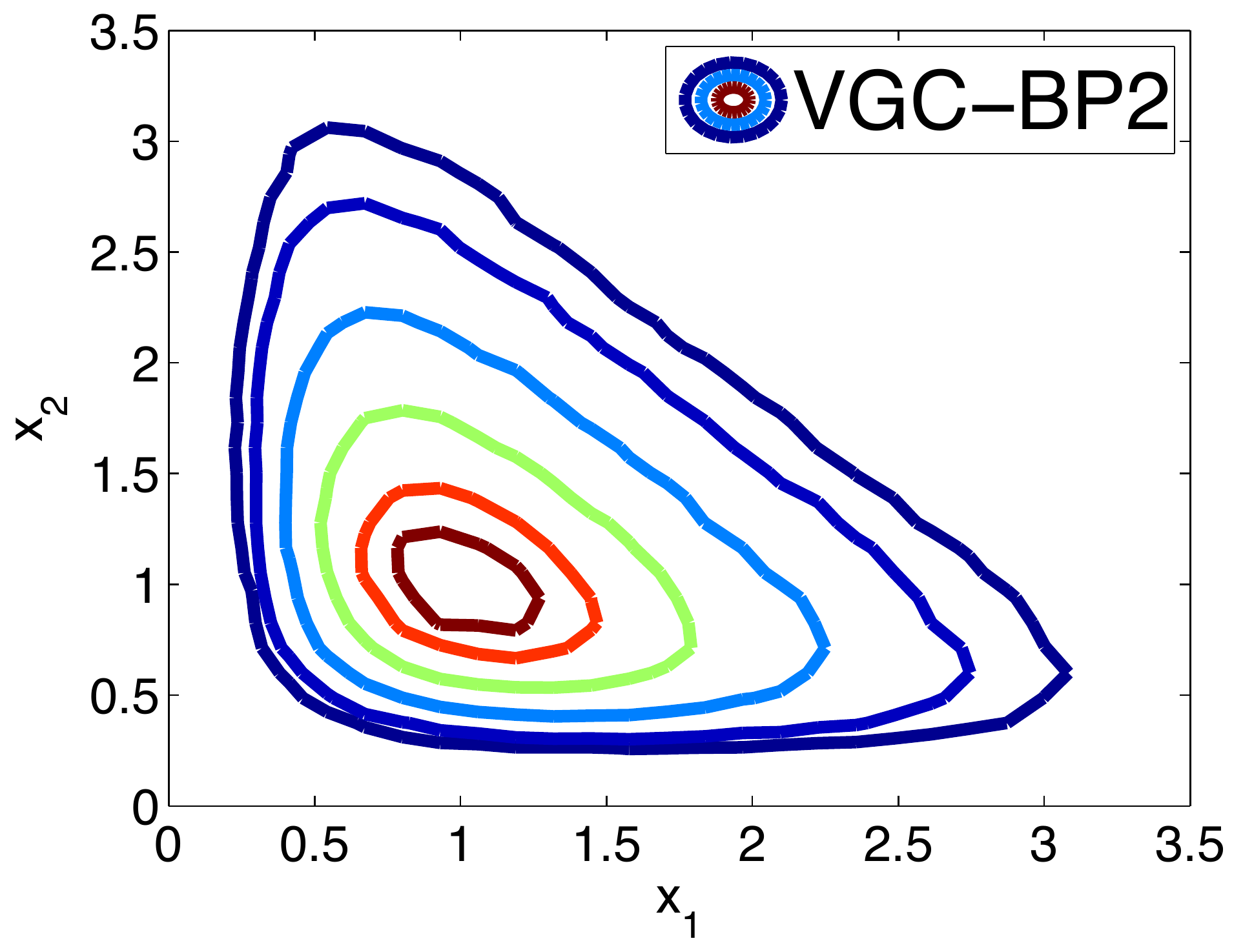}\\
\end{array}$}
\end{center}
\vskip -0.10in
\caption{Approximate Posteriors via VGC methods}
\label{figtwodLN} 
\end{figure}

We construct a bivariate Gaussian copula with (\rmnum{1}) Log-normal margins {(VGC-LN)} and (\rmnum{2}) BP-based margins {(VGC-BP)}. We set $\mu_{1}=\mu_{2}=0.1$ and $\sigma_{1}=\sigma_{2}=0.5$, $\rho = 0.4$ or $-0.4$ (first  and second row  in Figure \ref{figtwodLN}). Both VGC-LN and VGC-BP methods presume 
the correct form of the underlying copula (bivariate Gaussian) and learn the copula parameters $\rho$. VGC-LN further assumes exactly the true form of the univariate margins (log-normal) while VGC-BP is without any particular assumptions on parametric form of margins. Figure \ref{figtwodLN} shows that VGC-BP find as accurate joint posteriors as VGC-LN, even though the former assumes less knowledge about the true margins. 
  
\begin{figure}[bpht]
\vskip -0.07in
\begin{center}
\footnotesize{
$\begin{array}{cccc}
\hspace{-0.45cm}\includegraphics[height=0.12\textwidth, width=0.131\textwidth]{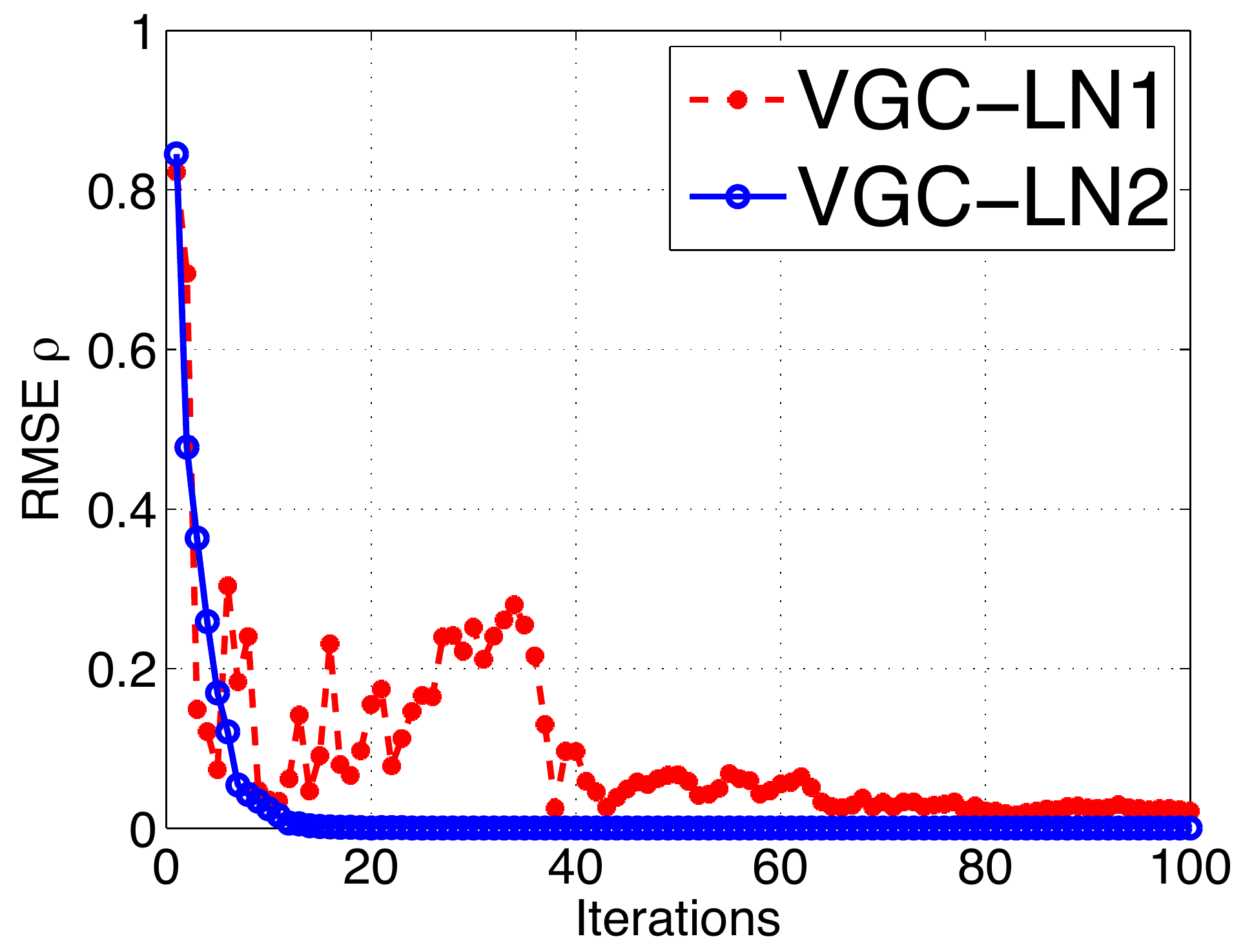} &
\hspace{-0.45cm}\includegraphics[height=0.12\textwidth, width=0.131\textwidth]{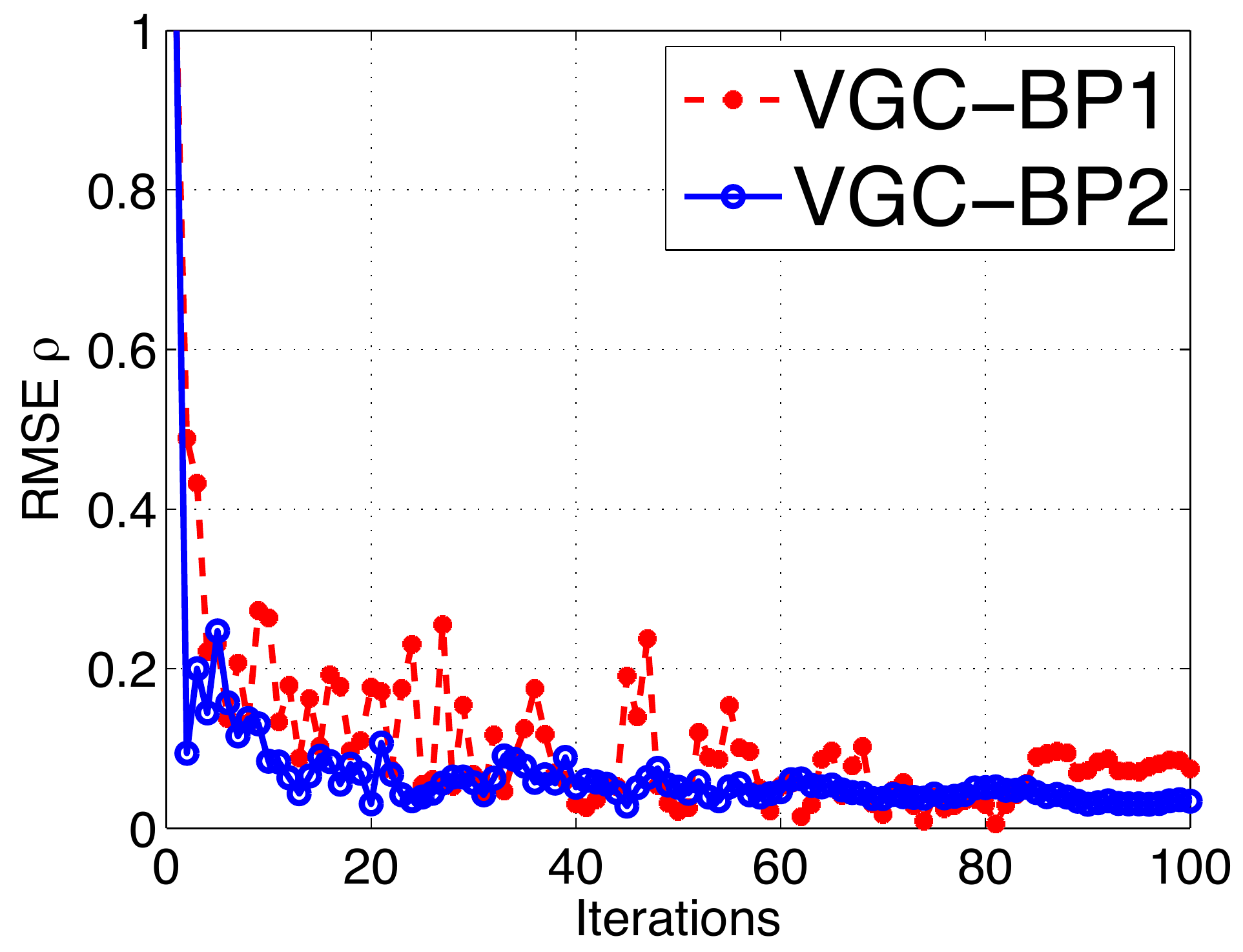} &
\hspace{-0.45cm}\includegraphics[height=0.12\textwidth, width=0.131\textwidth]{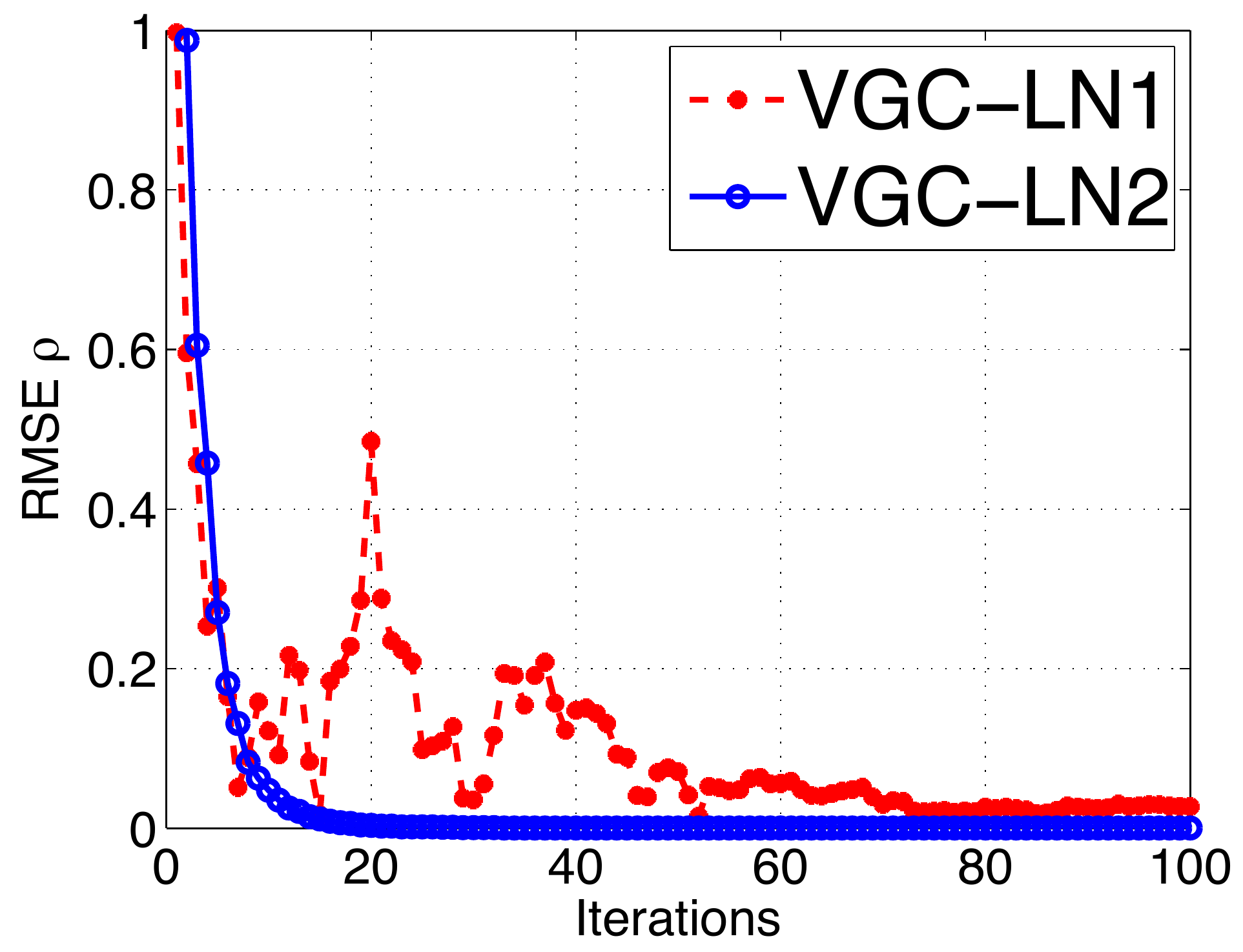} &
\hspace{-0.45cm}\includegraphics[height=0.12\textwidth, width=0.131\textwidth]{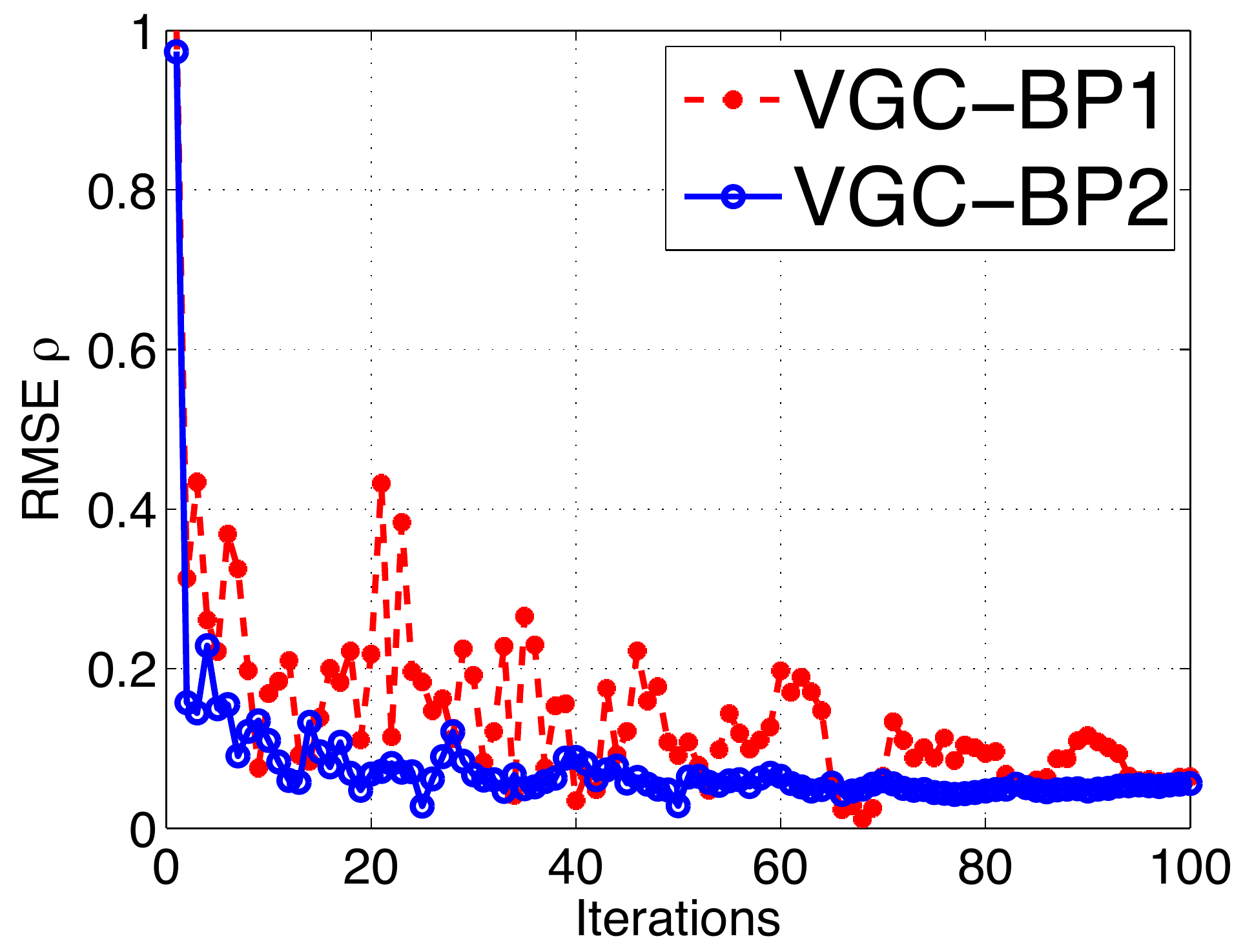}\\ 
\end{array}$}
\end{center}
\vskip -0.12in
\caption{ $\mathrm{RMSE}(\rho)$ of VGC-LN and VGC-BP v.s. Iterations; Left two: $\rho=0.4$; Right two: $\rho=-0.4$}
\label{figcompVG} 
\end{figure}

In updating $(\bd{\mu}, \bd{C})$, {VGC-LN1} and {VGC-BP1} follow the scheme in \citep{titsias2014doubly} and neglect the stochasticity in the entropy term; while {VGC-LN2} and {VGC-BP2} are based on our scheme in \eqref{ourscheme}.  Under the same learning rates, we define the relative mean square error (RMSE) of the copula parameter as ${R}(\rho)=\frac{(\hat{\rho}-\rho)^{2}}{\rho^2}$;   both VGC-LN and VGC-BP results  in Figure \ref{figcompVG} consistently show that our method leads to less noisy gradients and converges faster.   

%

 \subsection{Horseshoe Shrinkage}\label{secNIGG}
    
\begin{figure}
  \begin{minipage}[b]{0.65\linewidth}
    \centering
    {\small\begin{tabular}{c}
\hspace{-0.35cm}\includegraphics[width=5.2 cm,height=5.2cm]{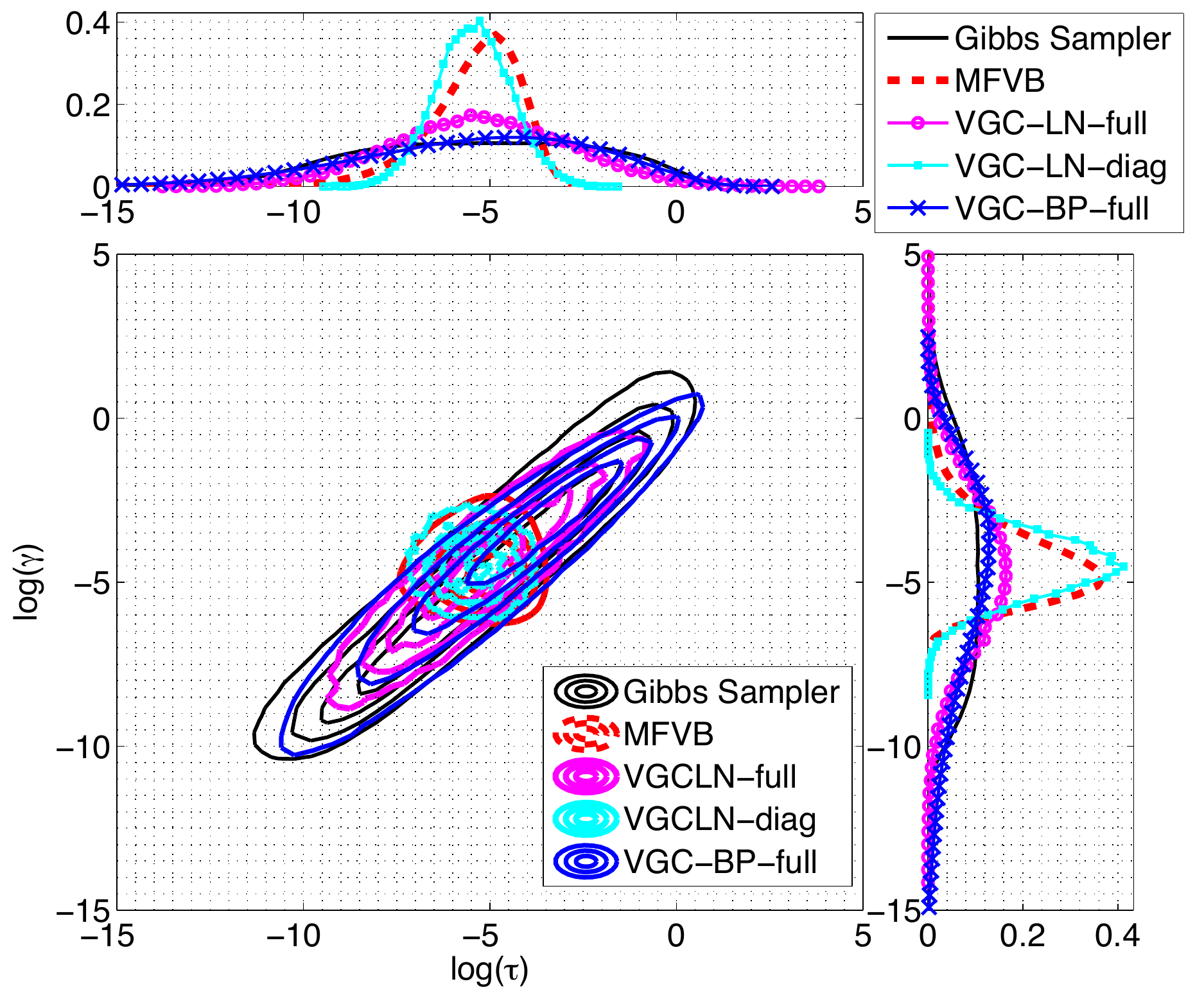}
\end{tabular}}
  \end{minipage}%
  \begin{minipage}[b]{0.3\linewidth}
    \centering%
    \scalebox{.7}{\begin{tabular}{|l|l|}
    \hline
Methods & ELBO \\ \hline
MFVB & -1.0778 \\ \hline
VGC-LN-full & -0.0634 \\ \hline
VGC-LN-diag & -1.2399 \\ \hline
VGC-BP &  0.3277\\ \hline
\end{tabular}}    \par\vspace{-4pt}
  \end{minipage}
  \vskip -0.12in
\caption{(Left Panel) Approximated Posteriors (Shown in Log Space for Visualization Purpose); (Right Panel) comparison of ELBO of different variational methods}
\vskip -0.15in
\label{fig1}
\end{figure}

The horseshoe distribution \citep{carvalho2010horseshoe} can be represented in equivalent conjugate hierarchies \citep{neville2014mean} $ y|\tau\sim \cN(0,\tau)$, $\tau|\lambda\sim \mathrm{InvGa}(0.5, \lambda)$, $ \lambda \sim \mathrm{InvGa}(0.5, 1)$. Here we assume $y=0.01$ is the (single) observation.  Denoting $\bd{x}=(x_{1}, x_{2})=(\tau, \gamma=1/\lambda)$, we implemented the VGC-BP algorithm ($k=10$) and VGC-LN algorithms (deterministic implementations\footnote{For gradient updates, we use a quasi-Newton strategy implemented in \cite{schmidt2012minfunc}.} are available in this special case). We compared them with two baselines: (\rmnum{1}) Gibbs sampler ($1\times 10^6$ samples), and (\rmnum{2}) MFVB. From Figure \ref{fig1}, it is noted that the VGC methods with full correlation matrix (VGC-LN-full, VGC-BP-full) are able to preserve the posterior dependence  and alleviate the under-estimation of the posterior variance. VGC-LN-full lead to higher ELBO than MFVB, and the gain is lost with factorized assumption $\bd{\Upsilon}=\bd{I}$ (VGC-LN-diag)  in which case the Gaussian copula reduces to the independence copula. The restriction of parametric margins is relaxed in VGC-BP.  With refinement of the mixture weights, VGC-BP leads to higher ELBO than VGC-LN. Since the Gaussian copula admits neither lower nor upper tail dependence, the posterior dependence it is able to preserve can be restrictive. It is a  future research topic  to explore other copula families  that allow more complex posterior dependencies in variational copula inference. 

 \subsection{Poisson Log-Linear Regression}\label{secPoLog}
\begin{figure}[bpht]
\vskip -0.02in
\begin{center}
\tiny{
$\begin{array}{cccc} 
\hspace{-0.35cm}\includegraphics[height=0.09\textwidth, width=0.12\textwidth]{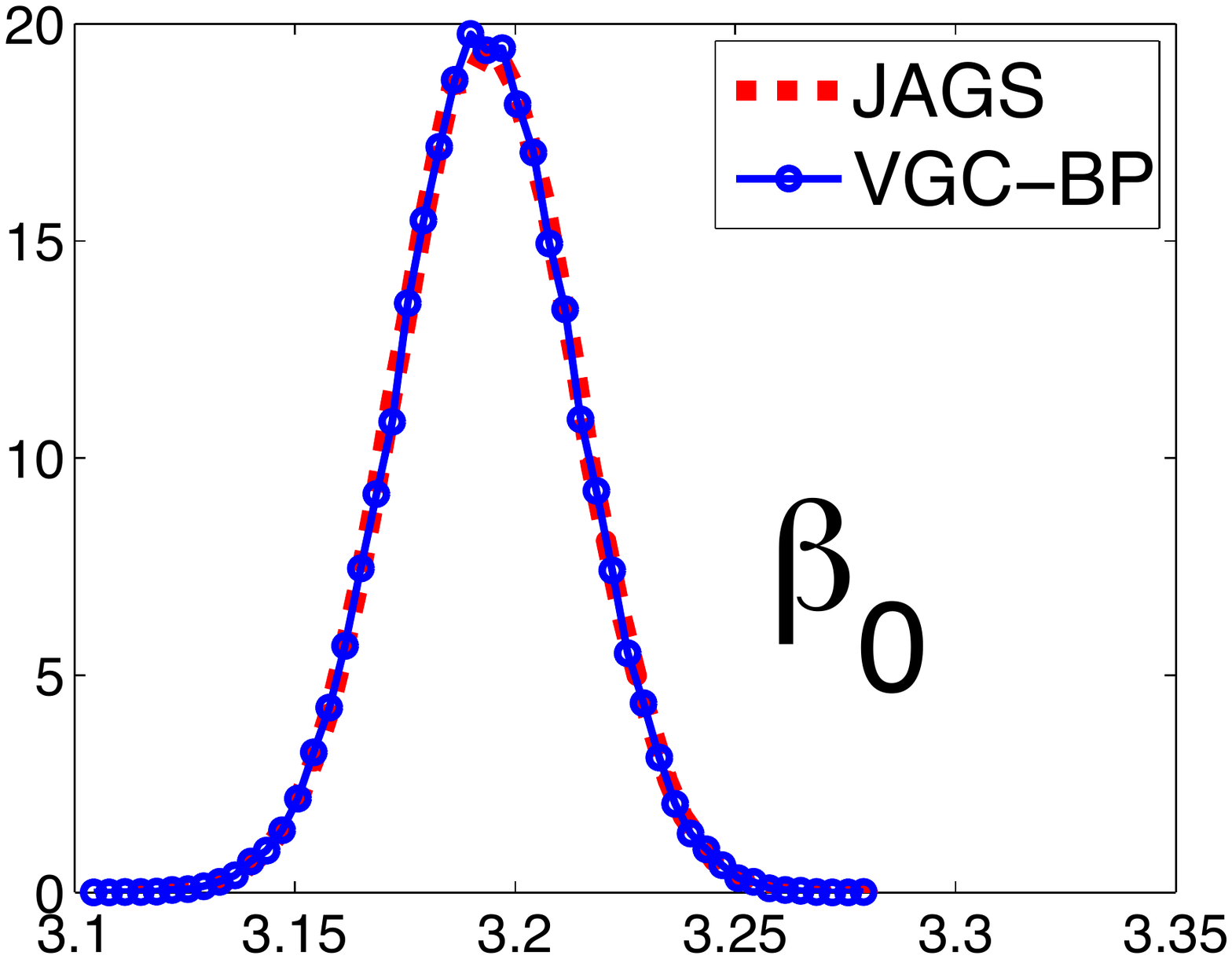} &
\hspace{-0.35cm}\includegraphics[height=0.09\textwidth, width=0.12\textwidth]{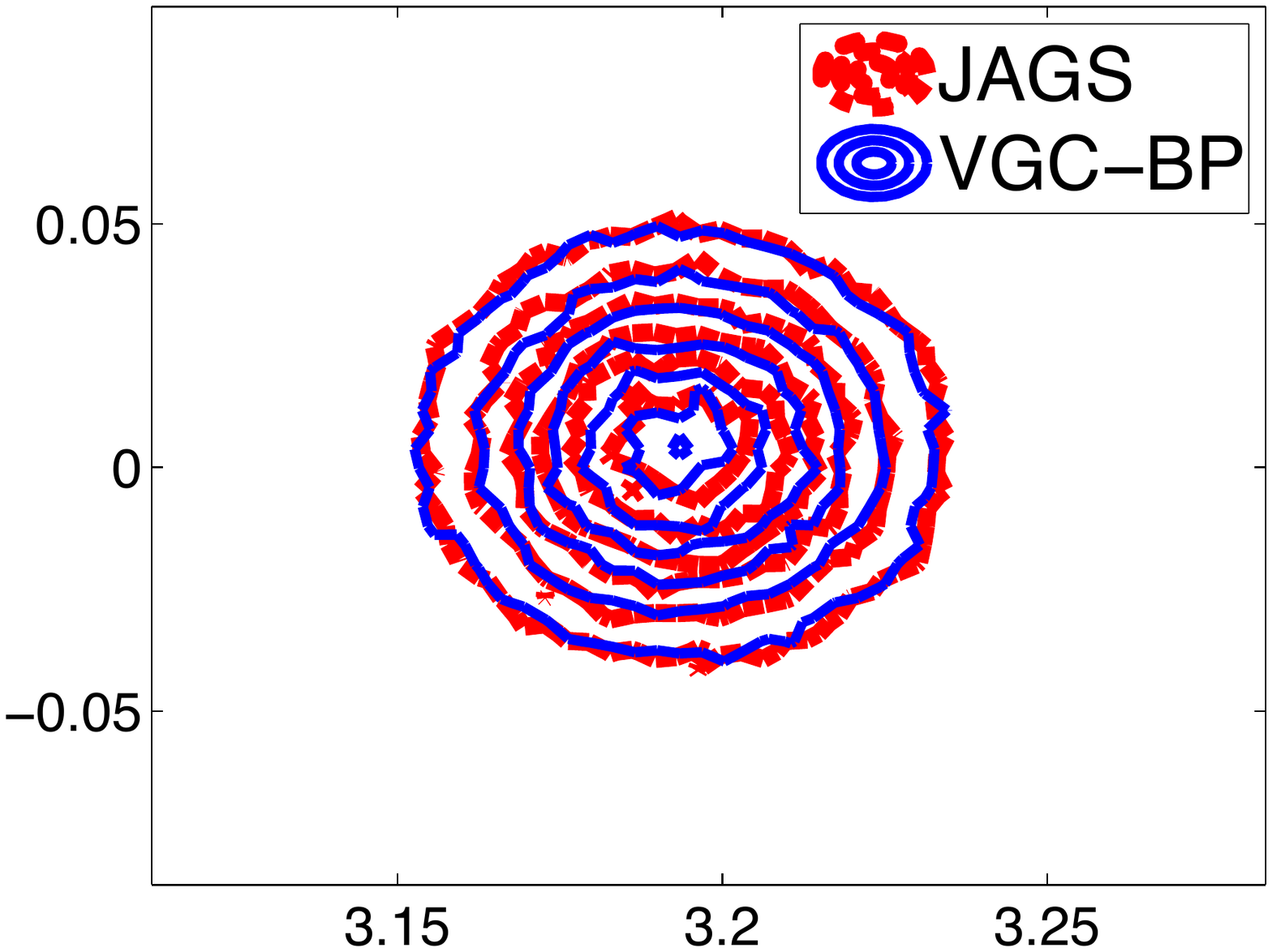} &
\hspace{-0.35cm}\includegraphics[height=0.09\textwidth, width=0.12\textwidth]{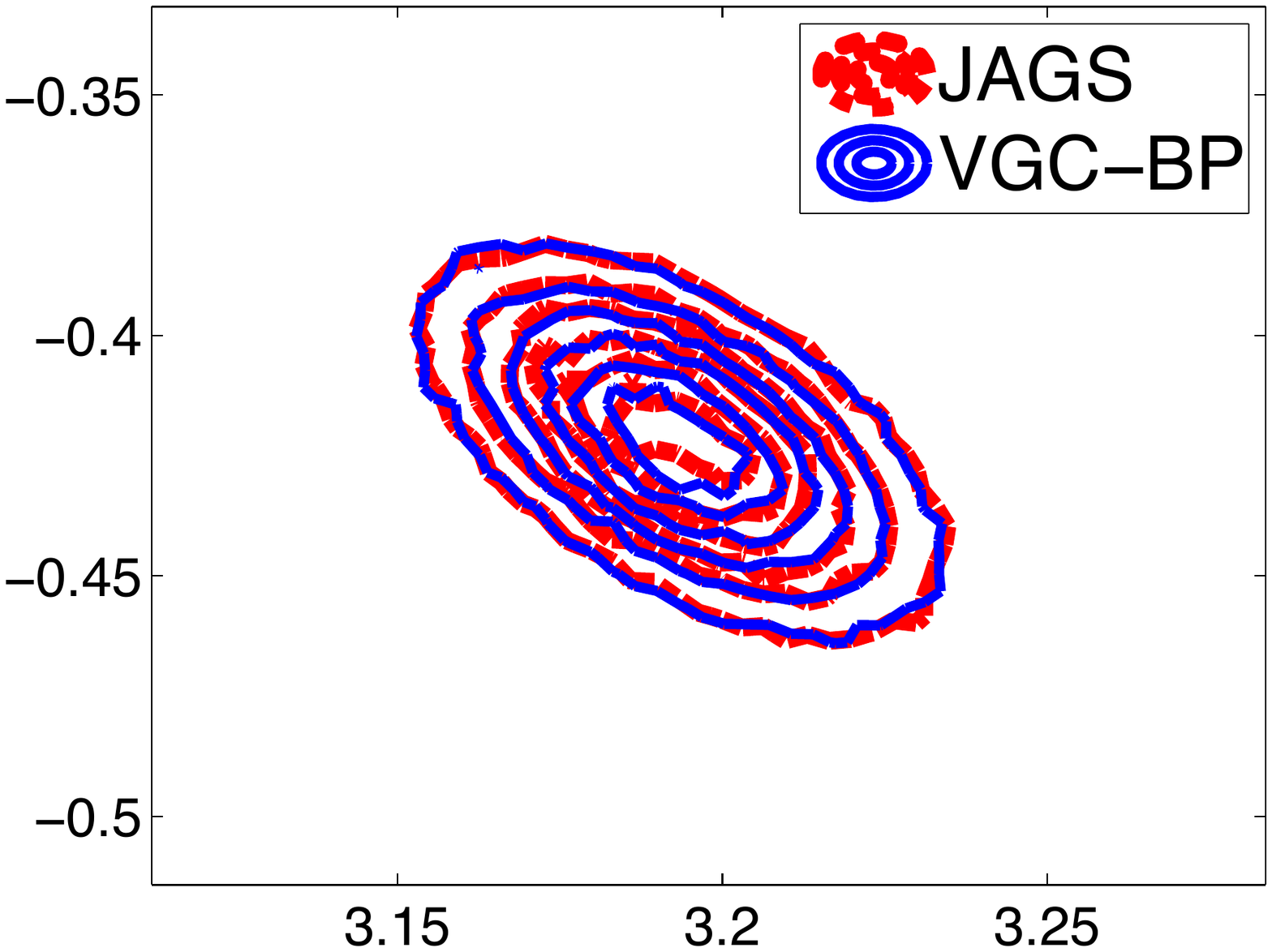} &
\hspace{-0.35cm}\includegraphics[height=0.09\textwidth, width=0.12\textwidth]{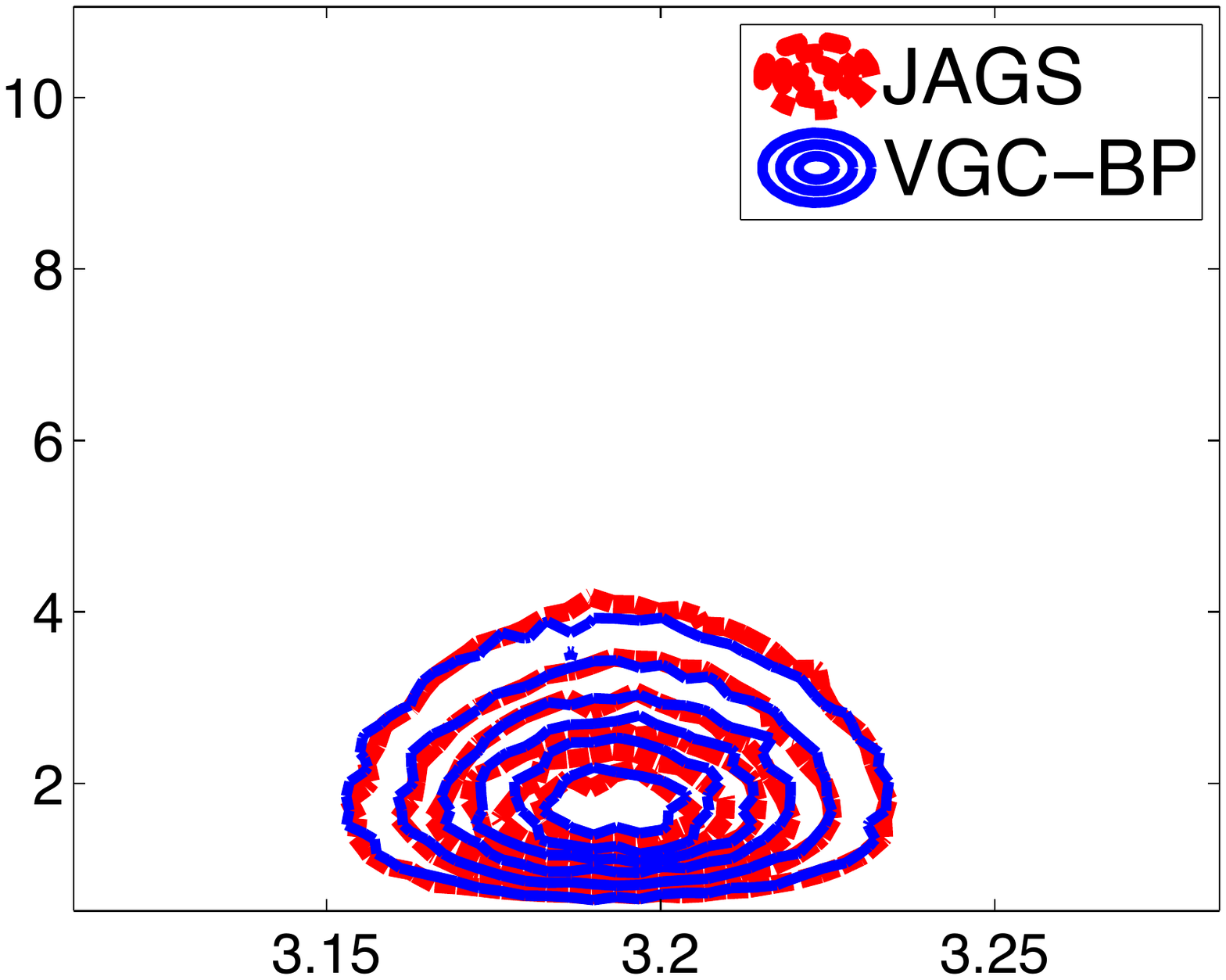}
\\[-0.3em] 
\hspace{-0.35cm}\includegraphics[height=0.09\textwidth, width=0.12\textwidth]{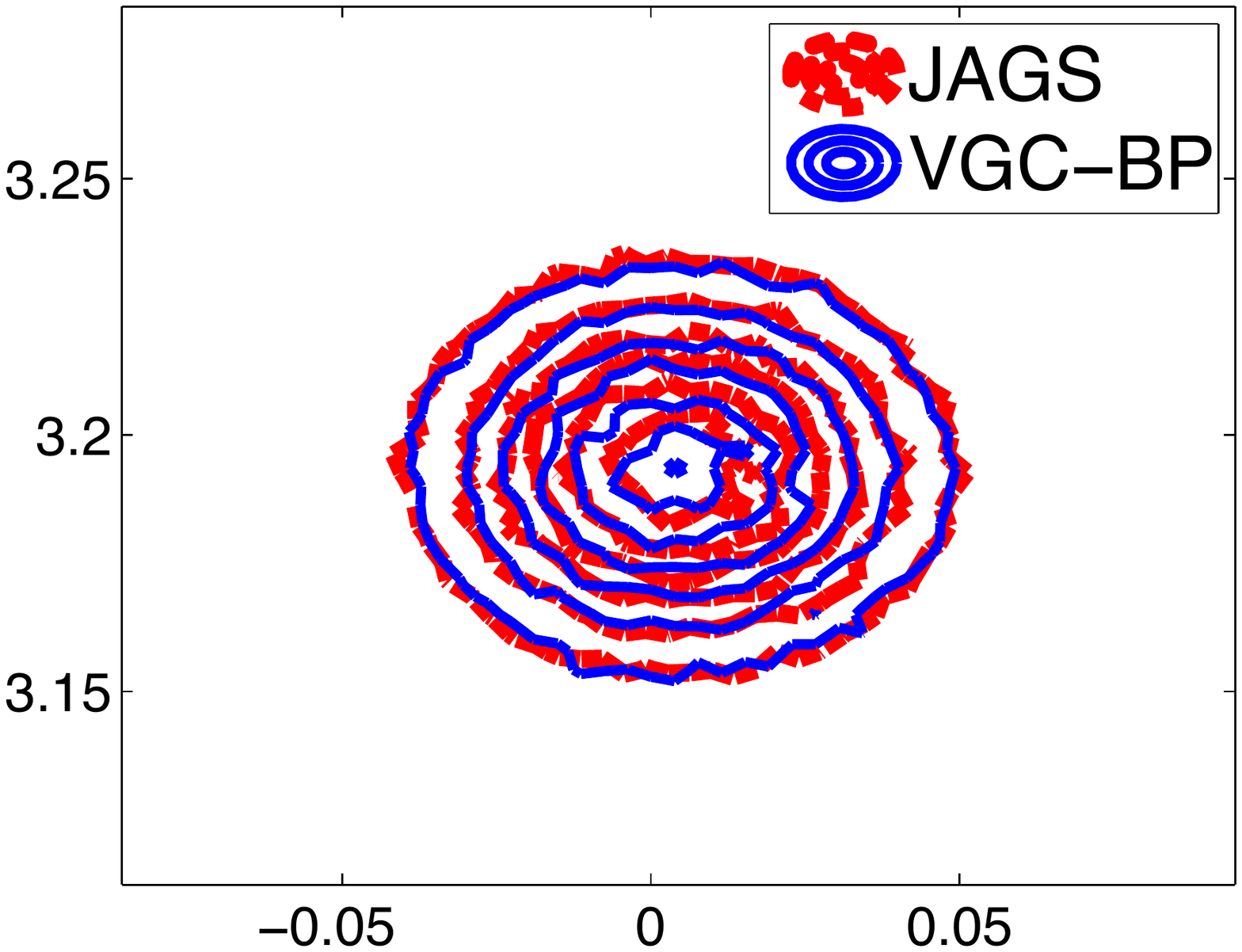} &
\hspace{-0.35cm}\includegraphics[height=0.09\textwidth, width=0.12\textwidth]{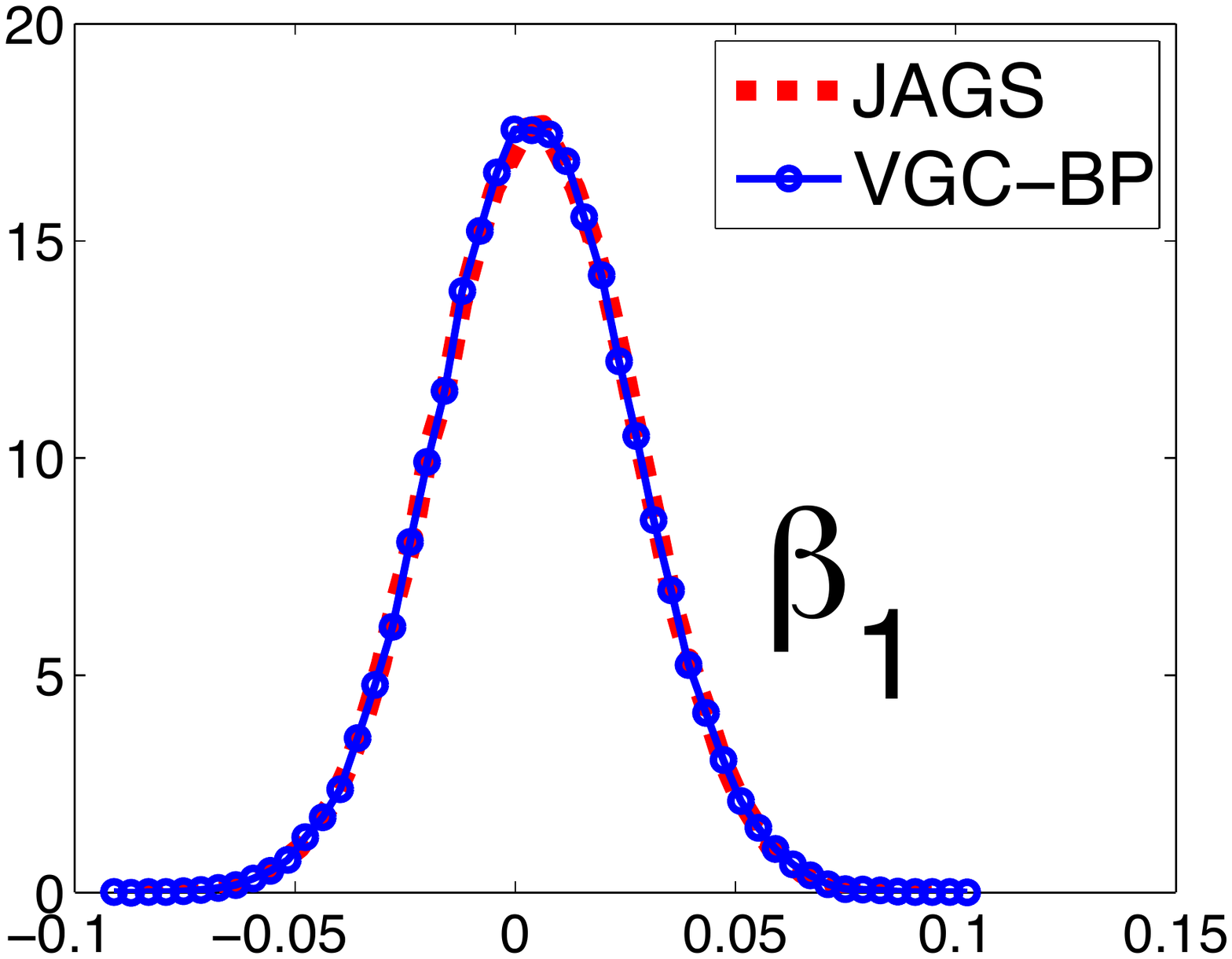} &
\hspace{-0.35cm}\includegraphics[height=0.09\textwidth, width=0.12\textwidth]{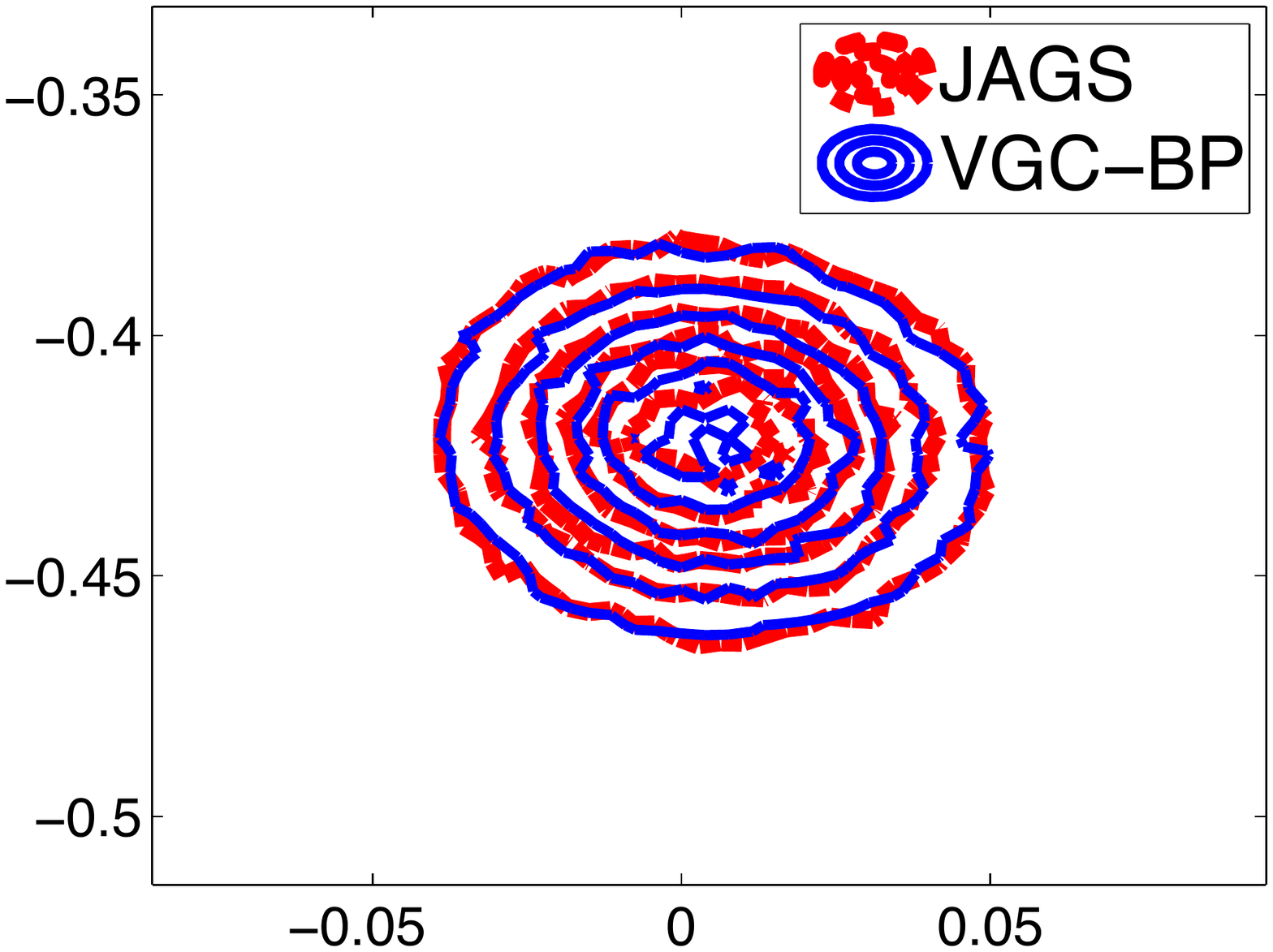} &
\hspace{-0.35cm}\includegraphics[height=0.09\textwidth, width=0.12\textwidth]{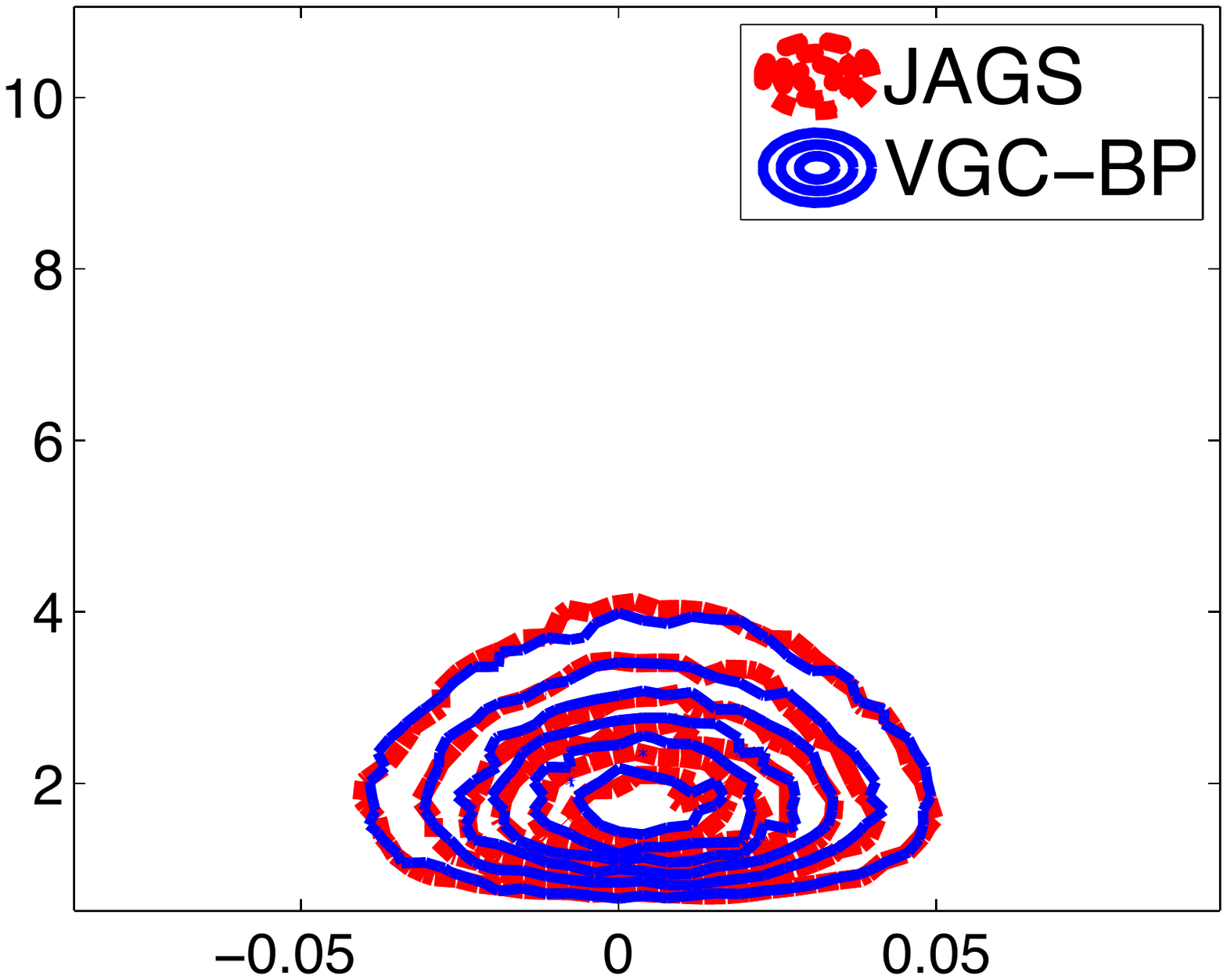}
\\[-0.3em] 
\hspace{-0.35cm}\includegraphics[height=0.09\textwidth, width=0.12\textwidth]{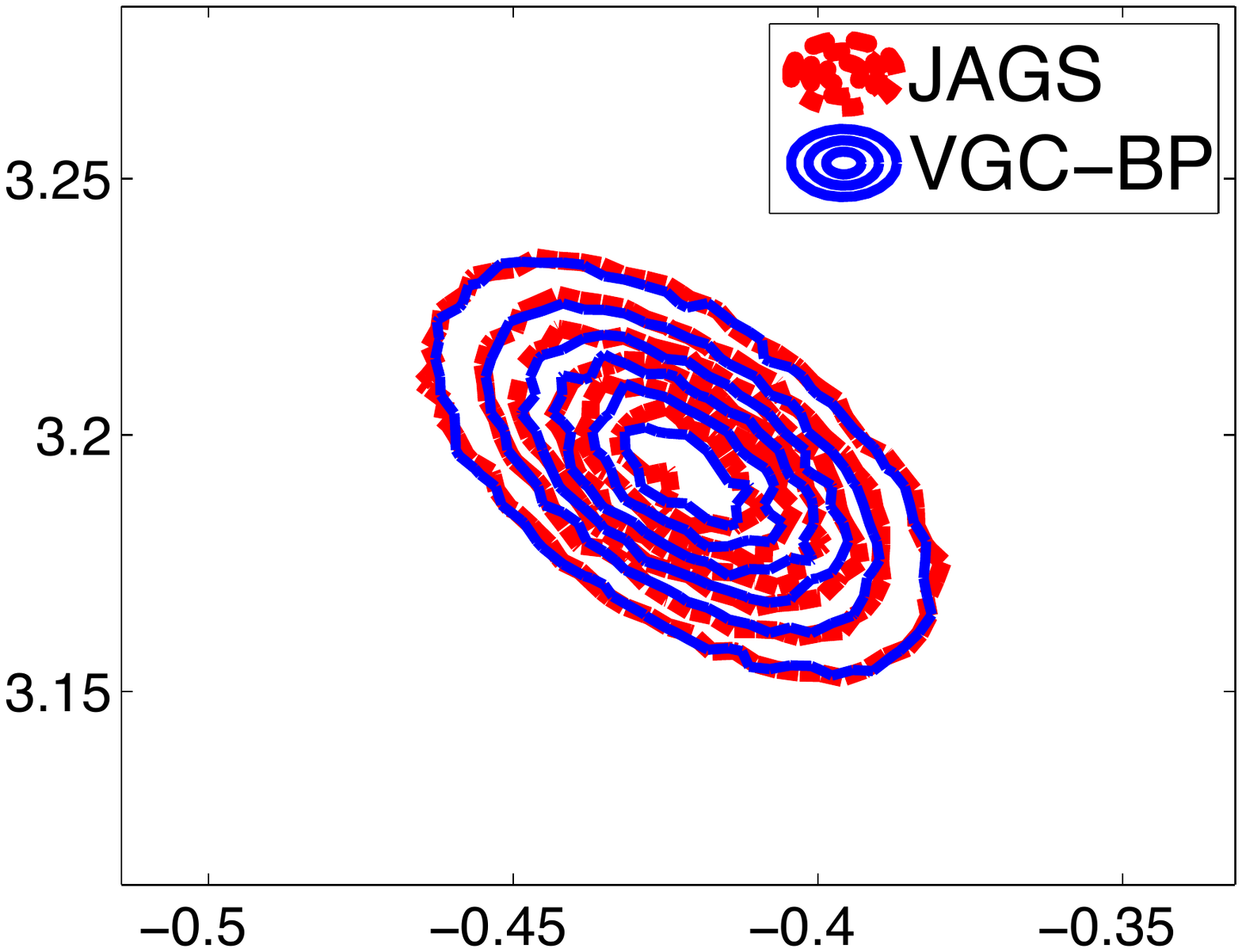} &
\hspace{-0.35cm}\includegraphics[height=0.09\textwidth, width=0.12\textwidth]{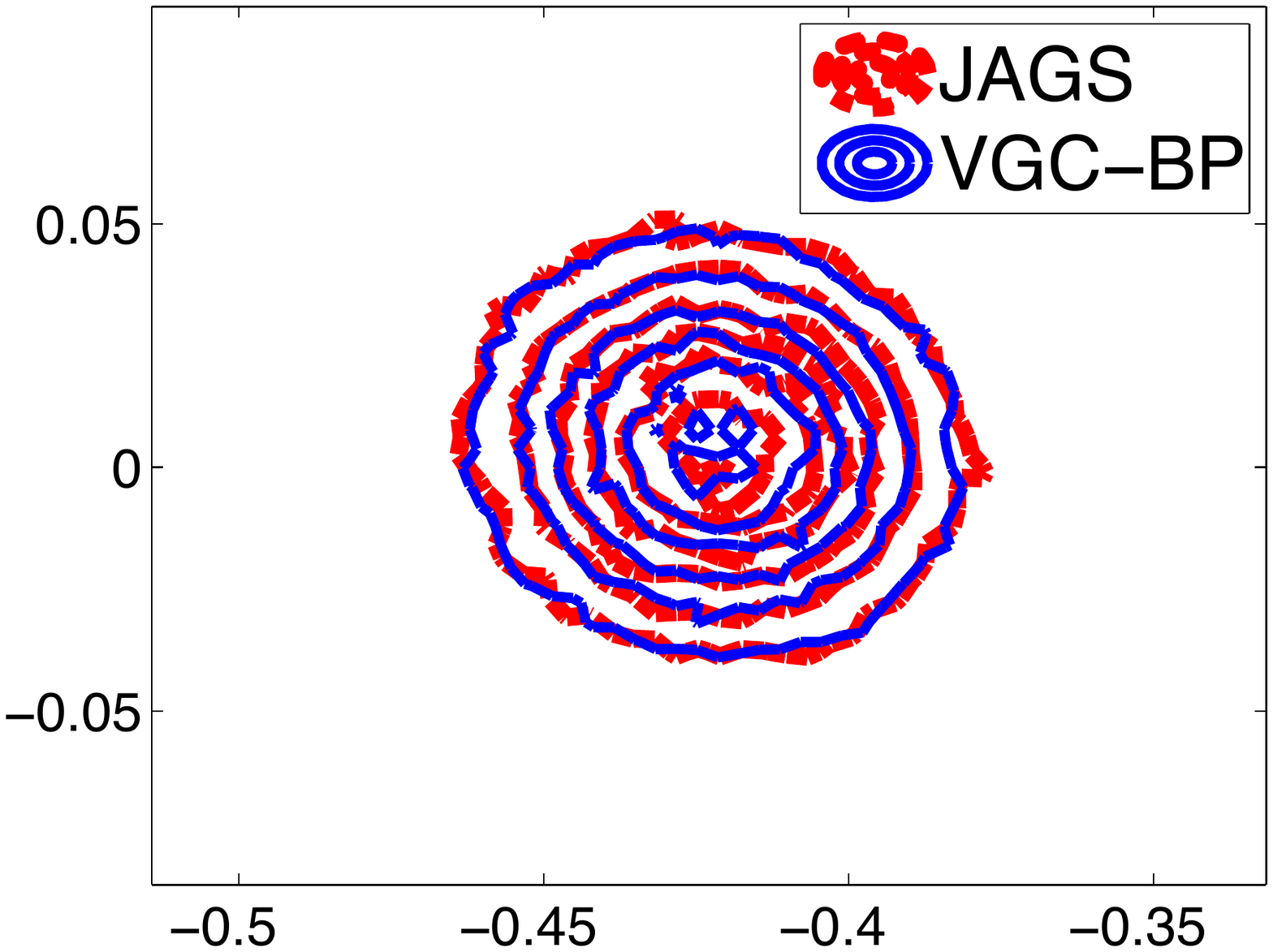} &
\hspace{-0.35cm}\includegraphics[height=0.09\textwidth, width=0.12\textwidth]{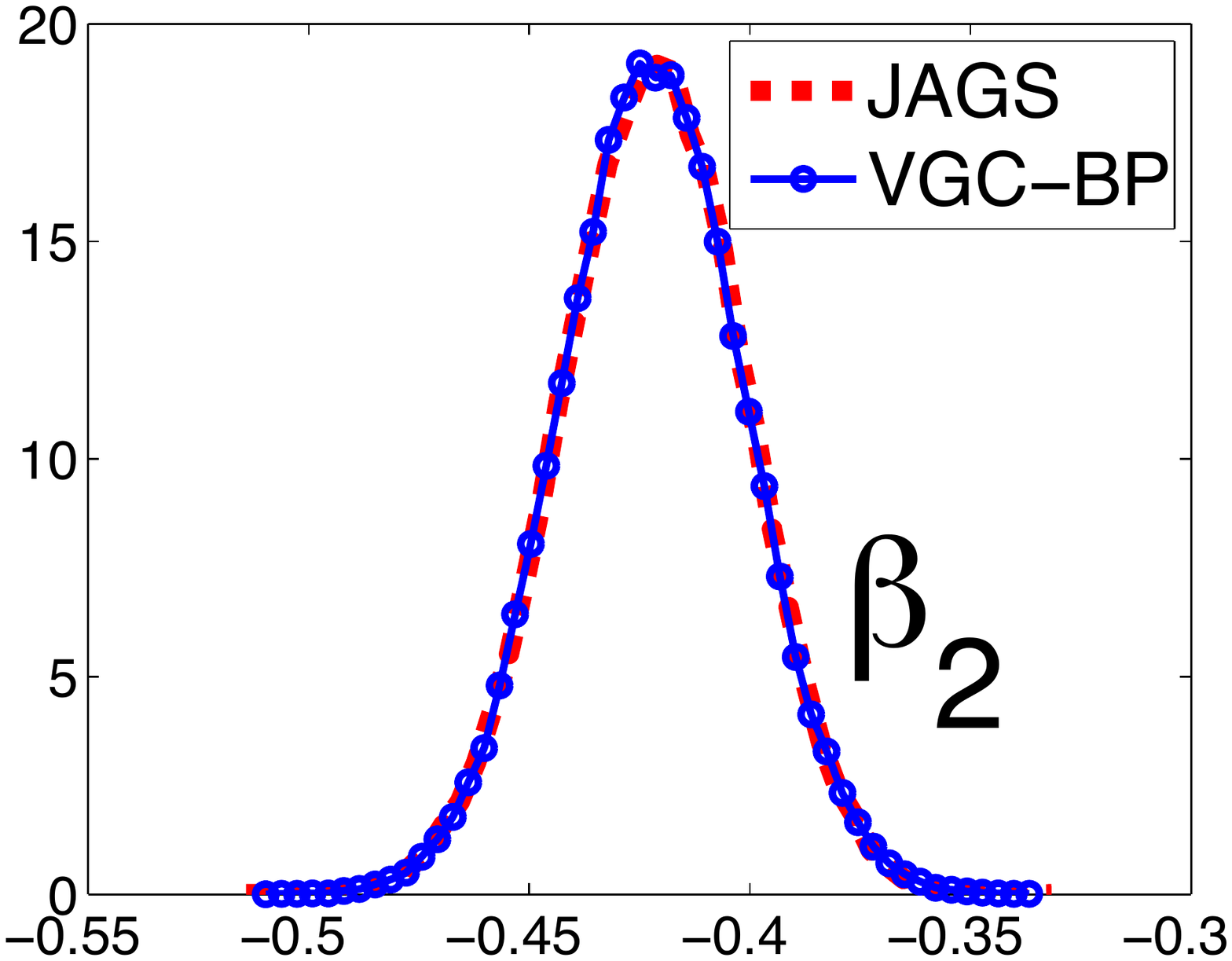} &
\hspace{-0.35cm}\includegraphics[height=0.09\textwidth, width=0.12\textwidth]{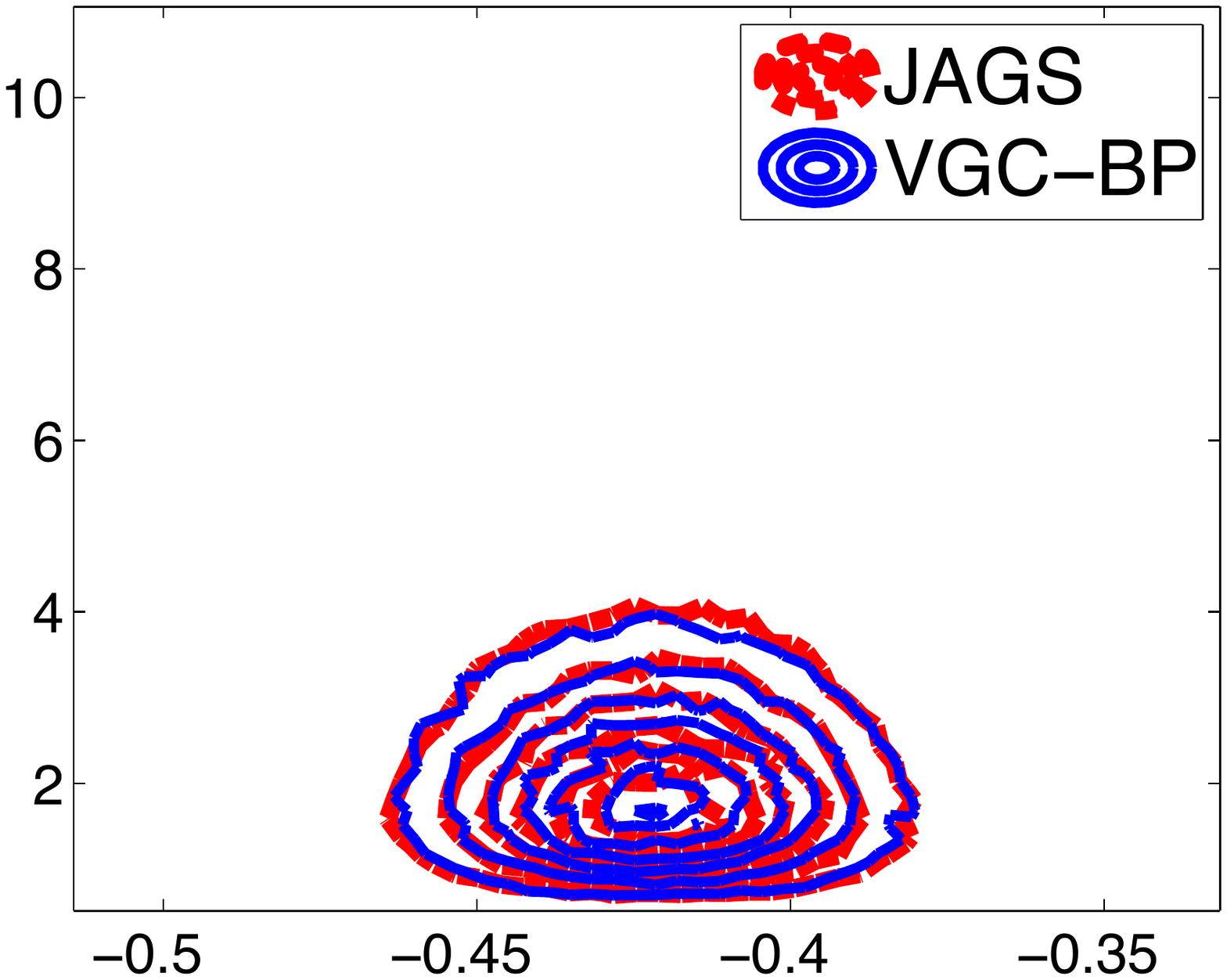}
\\[-0.3em] 
\hspace{-0.35cm}\includegraphics[height=0.09\textwidth, width=0.12\textwidth]{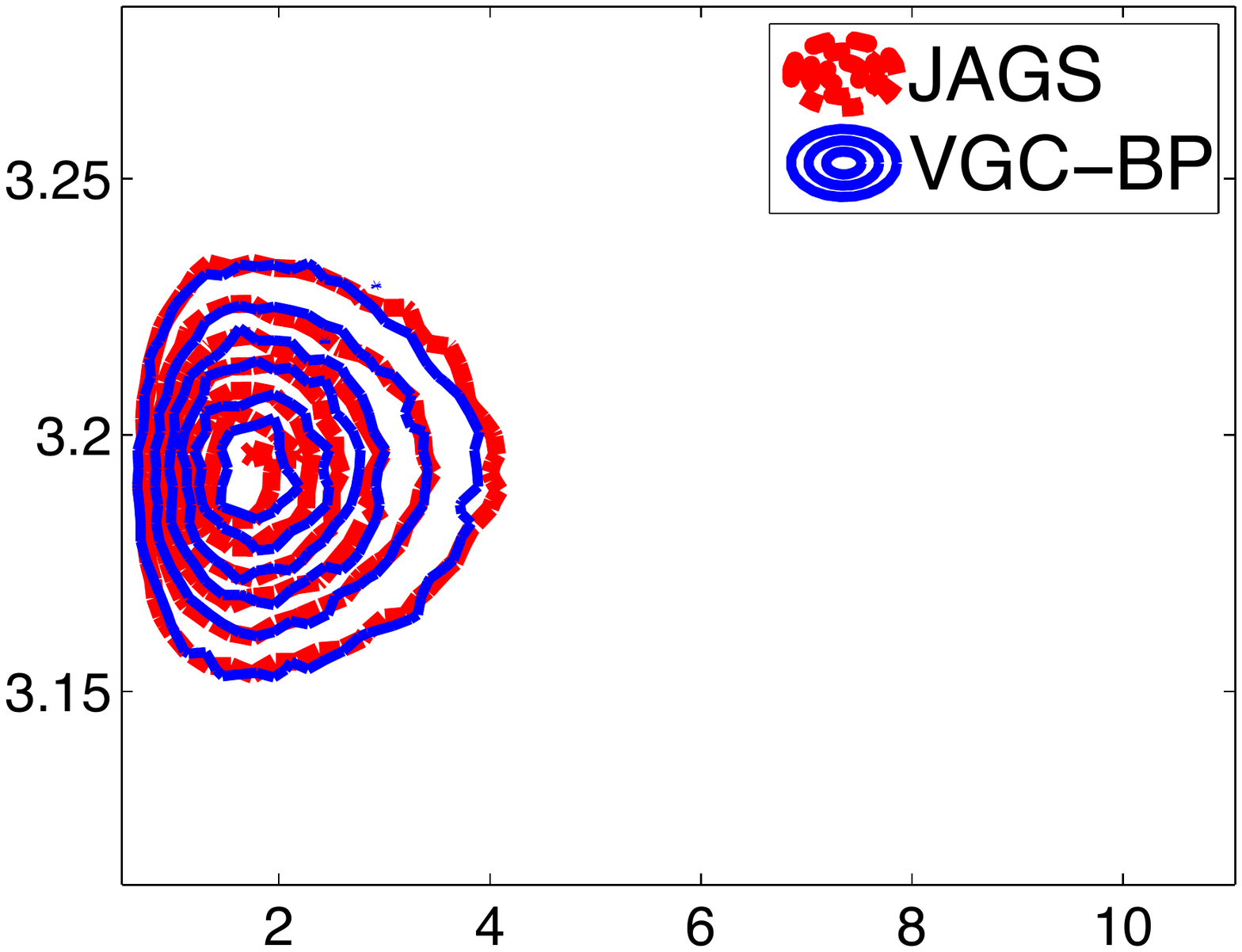} &
\hspace{-0.35cm}\includegraphics[height=0.09\textwidth, width=0.12\textwidth]{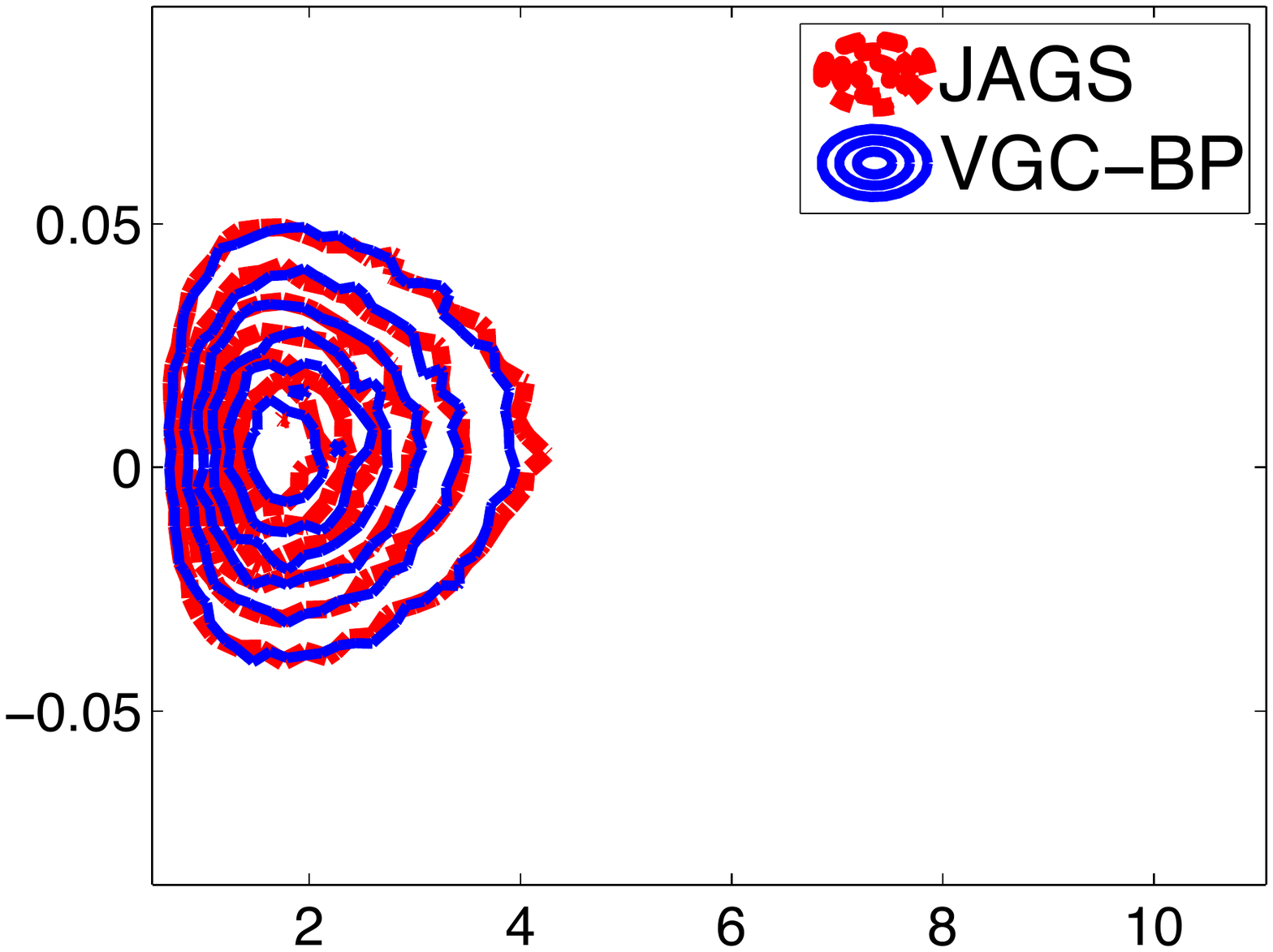} &
\hspace{-0.35cm}\includegraphics[height=0.09\textwidth, width=0.12\textwidth]{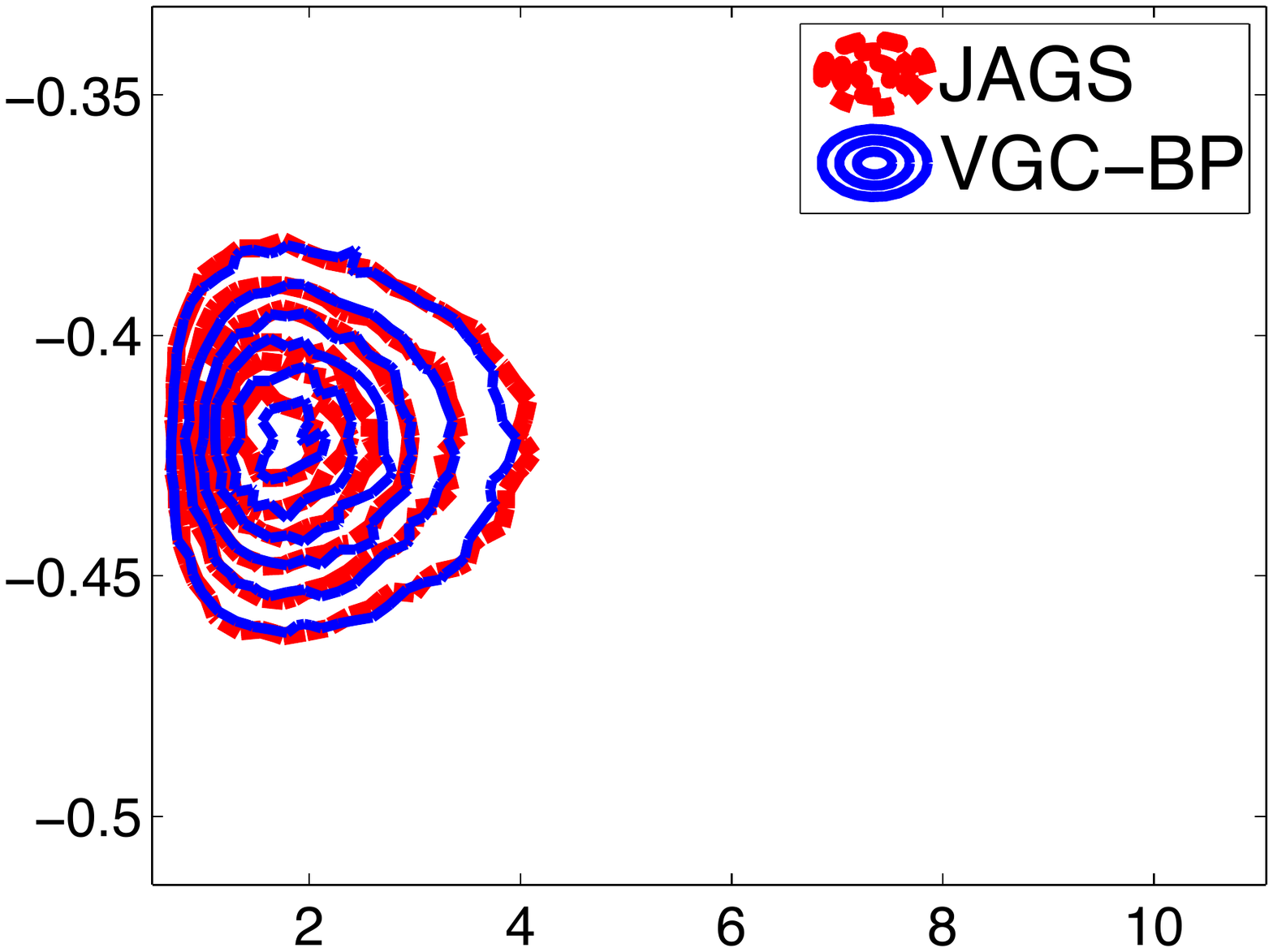} &
\hspace{-0.35cm}\includegraphics[height=0.09\textwidth, width=0.12\textwidth]{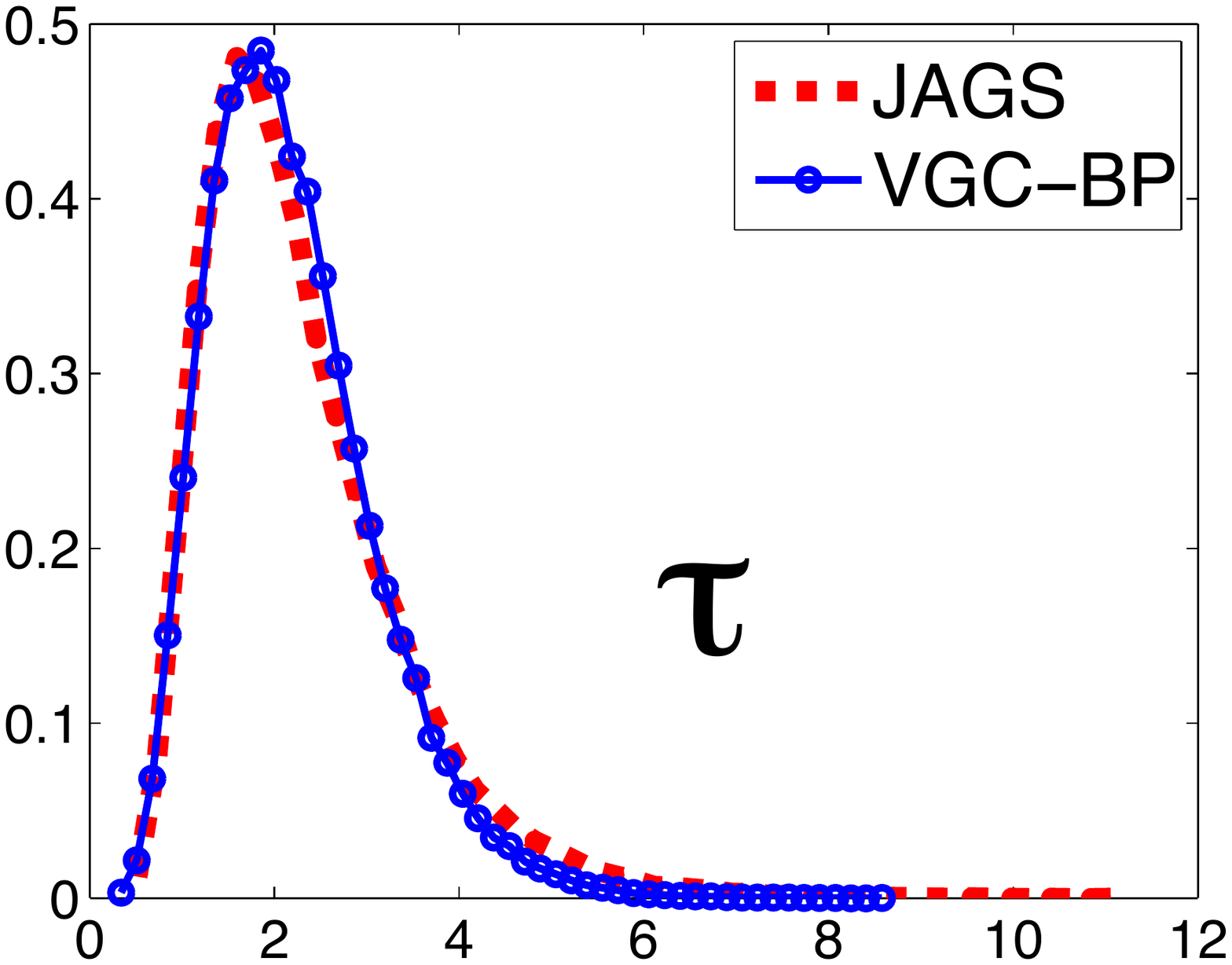}
\\[-0.3em] 
\end{array}$}
\end{center}
\vskip -0.12in
\caption{Univariate Margins and Pairwise Posteriors}
\vskip -0.15in
\label{figpolog} 
\end{figure}

We consider the tropical rain forest dataset \citep{moller2007modern}, a point pattern giving the locations of 3605 trees accompanied by covariate data  giving the elevation. Resampling the data into a grid of $50\times 50$m ($u_{i}$ locates the $i\mhyphen$th grid), the number of trees $y_{i}$ per unit area is modeled as, $y_{i}\sim \mathrm{Poisson}(\mu_{i})$, $i=1,\hdots, n$, $\log(\mu_{i})=\beta_{0} + \beta_{1}u_{i}+\beta_{2}u_{i}^2$,   $\beta_{0}\sim N(0, \tau)$, $\beta_{1}\sim N(0, \tau)$,  $\beta_{2}\sim N(0, \tau)$, $\tau\sim \mathrm{Ga}(1,1)$. We denote $\bd{x}=(\beta_{0}, \beta_{1}, \beta_{2}, \tau)$, and choosing $\Psi^{-1}(\cdot)$ to be  the CDF of $\cN(0,1)$ or $\mathrm{Exp}(1)$ accordingly. The implementation of VGC-BP leads to highly accurate marginal and pairwise posteriors  (See Figure \ref{figpolog}), as compared to the MCMC sampler ($1\times 10^{6}$ runs) implemented in JAGS \footnote{\url{http://mcmc-jags.sourceforge.net/}} as reference solutions. 

Interestingly, for non-conjugate models with unknown exact joint posteriors, VGC still provides a Sklar's representation of the approximated posterior,  including an analytical Gaussian copula, and a number of univariate margins (summarized  as univariate histograms if not in closed-form).  For further uses such as calculating sample quantiles,  simulating samples from $q_{\mathrm{VGC}}(\bd{x})$ is independent and faster, as compared to MCMC.  The obtained  posterior approximation could possibly improve the efficiency of  Metropolis-Hastings (MH) samplers by replacing the MCMC prerun as a reasonable proposal \citep{schmidl2013vine}.

The proposed method is an automated approach of approximating full posteriors. It is  readily applicable to a broad scope of latent Gaussian models with non-conjugate likelihoods. Compared with the integrated nested Laplace approximation (INLA) \citep{Rue09} and integrated non-factorized variational inference \citep{shaobo2013}, our approach does not need to discretize the space for non-Gaussian variables and thus does not suffer from the limits on the number of hyperparameters.

\vspace{-0.15cm}

\section{Discussions}

This article proposes a unified variational copula inference framework. In VGC, we have focused on Gaussian copula family for simplicity, however, other more flexible forms such as Gaussian mixture copula can be considered as well. To avoid the difficulty of specifying marginals for  hidden variables, a nonparametric procedure based on Bernstein polynomials indirectly induces highly flexible univariate margins. \cite{Dtran2015} and \cite{Kucukelbir2015} could potentially benefit from our flexible margins, while our approach is likely to benefit from the vine copula decomposition \citep{Dtran2015} to allow  richer or more complex dependencies and the automatic differentiation techniques applied in \cite{Kucukelbir2015}. 

\subsubsection*{Acknowledgments}
This research was supported in part by ARO, DARPA,
DOE, NGA, ONR and NSF.
%

\clearpage

\bibliography{Notes2015}
 
\newpage 

 \preto\align{\par\nobreak\small\noindent}

\section*{Supplementary Material}
\subsection*{A:  KL  Additive Decomposition}
Letting the variational proposal in Sklar's representation be $q_{\mathrm{VC}}(\bd{x})=c(\bd{u})\prod_{j=1}^{p}f_{j}(x_{j})$, and the true posterior be $p(\bd{x}|\bd{y})=c^{\star}(\bd{v})\prod_{j}f_{j}^{\star}(x_{j})$, where $\bd{u}=F(\bd{x})=[F_{1}(x_{1}),\hdots, F_{p}(x_{p})]$, $\bd{v}=F^{\star}(\bd{x})=[F_{1}^{\star}(x_{1}),\hdots, F_{p}^{\star}(x_{p})]$.  The KL divergence decomposes into additive terms,  
\begin{align}\label{KLdecompose}
&\mathrm{KL}\{q(\bd{x})||p(\bd{x}|\bd{y})\}=\int q(\bd{x})\bigg(\log{\frac{q(\bd{x})}{p(\bd{x}|\bd{y})}}\bigg)d\bd{x}\cr 
&=\int c[F(\bd{x})]\prod_{j}f_{j}(x_{j})\bigg(\log{\frac{c[F(\bd{x})]\prod_{j}f_{j}(x_{j})}{c^{\star}[F^{\star}(\bd{x})]\prod_{j}f_{j}^{\star}(x_{j})}}\bigg)d\bd{x}\cr 
&=\int c[F(\bd{x})]\bigg(\log{\frac{c[F(\bd{x})]}{c^{\star}[F^{\star}(\bd{x})]}}\bigg)\prod_{j}d F_{j}({x}_{j})\cr &+\int c[F(\bd{x})]\prod_{j}f_{j}(x_{j})\bigg(\log{\frac{\prod_{j}f_{j}(x_{j})}{\prod_{j}f_{j}^{\star}(x_{j})}}\bigg)\prod_{j}d {x}_{j}.
\end{align}
The first term in \eqref{KLdecompose}
\begin{align}
&\int c[F(\bd{x})]\bigg(\log{\frac{c[F(\bd{x})]}{c^{\star}[F^{\star}(\bd{x})]}}\bigg)\prod_{j}d F_{j}({x}_{j})\cr 
&=\int c(\bd{u})\bigg(\log{\frac{c(\bd{u})}{c^{\star}(F^{\star}(F^{-1}(\bd{u})))}}\bigg) d \bd{u}\cr
&=\mathrm{KL}\{c(\bd{u})||c^{\star}[F^{\star}(F^{-1}(\bd{u}))]\},\nonumber
\end{align}
The second term in \eqref{KLdecompose}
\begin{align}
&\int c[F(\bd{x})]\prod_{j}f_{j}(x_{j})\bigg(\log{\frac{\prod_{j}f_{j}(x_{j})}{\prod_{j}f_{j}^{\star}(x_{j})}}\bigg)\prod_{j}d {x}_{j}\cr 
&=\sum_{j}\int c[F(\bd{x})]\prod_{j}f_{j}(x_{j})\bigg(\log{\frac{f_{j}(x_{j})}{f_{j}^{\star}(x_{j})}}\bigg)\prod_{j}d {x}_{j}\cr
&=\sum_{j}\int f_{j}(x_{j})\bigg(\log{\frac{f_{j}(x_{j})}{f_{j}^{\star}(x_{j})}}\bigg)d {x}_{j}~ \textrm{\scriptsize{(Marginal Closed Property)}}\cr
&=\sum_{j}\mathrm{KL}\{f_{j}(x_{j})||f_{j}^{\star}(x_{j})\},\nonumber
\end{align}
Therefore
\begin{align}
\mathrm{KL}\{q(\bd{x})||p(\bd{x}|\bd{y})\}&=\mathrm{KL}\{c[F(\bd{x})]||c^{\star}[F^{\star}(\bd{x})]\}\cr &+\sum_{j}\mathrm{KL}\{f_{j}(x_{j})||f_{j}^{\star}(x_{j})\}
\end{align}
\subsection*{B:  Model-Specific Derivations}
\subsubsection*{B1: Skew Normal Distribution}
\begin{enumerate}
\item   $\ln p(x)\propto \ln{\phi(x)}+\ln{\Phi({\alpha}x)}$
and ${\partial \ln p(x)}/{\partial x}=-x+{\alpha\phi(\alpha x)}/{\Phi(\alpha x)}$, $\alpha$ is the shape parameter
\item  $\Psi(x)$ is predefined as CDF of $\cN(0, 1)$
\end{enumerate}
\subsubsection*{B2: Student's t Distribution}
\begin{enumerate}
\item   $\ln p(x)\propto -{(\nu+1)}/{2}\ln{(1+{x^2}/{\nu})}$ and ${\partial \ln p(x)}/{\partial x}=-{(\nu+1)x}/{(\nu+x^2)}$, $\nu>0$ is the degrees of freedom 
\item  $\Psi(x)$ is predefined as CDF of $\cN(0, 1)$
\end{enumerate}

\subsubsection*{B3: Gamma Distribution}
\begin{enumerate}
\item   $\ln p(x)\propto(\alpha-1)\ln{x}-\beta x$ 
and ${\partial \ln p(x)}/{\partial x}={(\alpha-1)}/{x}-\beta$, $\alpha$ is the shape parameter, $\beta$ is the rate parameter 
\item  $\Psi(x)$ is predefined as CDF of   $\mathrm{Exp}(1)$ 
\end{enumerate}

\subsubsection*{B4: Beta Distribution}
\begin{enumerate}
\item   $\ln p(x)\propto(a-1)\ln{x}+(b-1)\ln{(1-x)}$ 
and ${\partial \ln p(x)}/{\partial x}={(a-1)}/{x}-{(b-1)}/{(1-x)}$, both $a$, $b>0$
\item  $\Psi(x)$ is predefined as CDF of $\mathrm{Beta}(2,2)$ 
\end{enumerate}

\subsubsection*{B5: Bivariate Log-Normal}
%
\begin{enumerate}
\item   $\ln p(x_{1}, x_{2})\propto -\ln{x_{1}}-\ln{x_{2}}-{\zeta}/{2}$ 
and \begin{align}
\frac{\partial \ln f(x_{1}, x_{2})}{\partial x_{1}} &= -\frac{1}{x_{1}}-\frac{\alpha_{1}(x_{1})-\rho \alpha_{2}(x_{2})}{(1-\rho^2)x_{1}\sigma_{1}}\cr 
\frac{\partial \ln f(x_{1}, x_{2})}{\partial x_{2}} &= -\frac{1}{x_{2}}-\frac{\alpha_{2}(x_{2})-\rho \alpha_{1}(x_{1})}{(1-\rho^2)x_{2}\sigma_{2}} \nonumber
\end{align}
\item  $\Psi(x)$ is predefined as CDF of $\mathrm{Exp}(1)$  
\end{enumerate}

\subsection*{C. Derivations in Horseshoe Shrinkage}
The equilvalent hierarchical model is 
 \begin{align}
 y|\tau\sim \cN(0,\tau), \quad \tau|\gamma\sim \mathrm{InvGa}(0.5, \gamma),\quad \gamma \sim \mathrm{Ga}(0.5, 1)\nonumber
 \end{align}
\subsubsection*{C1: Gibbs Sampler}
The full conditional posterior distributions are 
  \begin{align}
p(\tau|y, \gamma)=\mathrm{InvGa}\left(1, {y^2}/{2}+\gamma\right),\quad p(\gamma|\tau)=\mathrm{Ga}\left(1, {\tau}^{-1}+1\right)\nonumber
   \end{align}

%
%
   
 \subsubsection*{C2: Mean-field Variational Bayes}
  The ELBO under MFVB is 
  \begin{align}
\mathcal{L}_{\mathrm{MFVB}}[q_{\mathrm{VB}}(\tau,\gamma)]&=\mathbb{E}_{q(\tau)q(\gamma)}[\ln p(y, \tau, \gamma)]\cr &+H_{1}[q(\tau; \alpha_{1}, \beta_{1})]+H_{2}[q(\gamma; \alpha_{2}, \beta_{2})]\nonumber
 \end{align}
where 
\begin{align}
&\mathbb{E}_{q(\tau)q(\gamma)}[\ln p(y, \tau, \gamma)]=-0.5\ln{(2\pi)}-2\ln{\Gamma(0.5)}-2\langle\ln{\tau}\rangle\cr &-{y^2}\left\langle{\tau}^{-1}\right\rangle/2-\langle\gamma\rangle\left\langle{\tau}^{-1}\right\rangle-\langle\gamma\rangle\cr
& H_{1}[q(\tau; \alpha_{1}, \beta_{1})] = \alpha_{1}+\ln{\beta_{1}}+\ln{[\Gamma(\alpha_{1})]}-(1+\alpha_{1})\psi(\alpha_{1}) \cr 
& H_{2}[q(\gamma; \alpha_{2}, \beta_{2})] = \alpha_{2}-\ln{\beta_{2}}+\ln{[\Gamma(\alpha_{2})]}+(1-\alpha_{2})\psi(\alpha_{2})\nonumber
  \end{align}
The variational distribution 
 \begin{align}
 q(\tau)&=\mathcal{I}\cG\left(\tau; \alpha_{1}, \beta_{1}\right)=\mathcal{I}\cG\left(\tau; 1, {y^2}/{2}+\langle\gamma\rangle\right),\cr 
 q(\gamma)&=\cG(\gamma; \alpha_{2}, \beta_{2})=\cG\left(\gamma; 1, \left\langle{\tau}^{-1}\right\rangle+1\right)\nonumber
 \end{align}
 where 
  \begin{align}
\langle\ln{\tau}\rangle &=\ln{\beta_{1}}-\psi(\alpha_{1})=\ln{\left({y^2}/{2}+\langle\gamma\rangle\right)}-\psi(1),\cr
\left\langle{\tau}^{-1}\right\rangle&=\frac{\alpha_{1}}{\beta_{1}}=\frac{1}{\left({y^2}/{2}+\langle\gamma\rangle\right)},\quad
\langle\gamma\rangle=\frac{\alpha_{2}}{\beta_{2}}=\frac{1}{\left\langle{\tau}^{-1}\right\rangle+1}\nonumber
  \end{align}

\subsubsection*{C3: Deterministic VGC-LN}

Denoting $\bd{x}=(x_{1}, x_{2})=(\tau, \gamma)$, we construct a variational Gaussian copula  proposal with  (1) a bivariate Gaussian copula, and (2) fixed-form margin for both {\small$x_{1}=\tau\in(0, \infty)$} and {\small$x_{2}=\gamma\in(0, \infty)$}; we employ   
{$f_{j}(x_{j}; \mu_{j}, \sigma_{jj}^{2})={\mathcal{L}\cN(x_{j}; \mu_{j},\sigma_{jj}^{2})}$},{$x_{j}=h_{j}(\wt{z_{j}})=\exp{(\wt{z_{j}})}=g(z_{j})=\exp{({\sigma_{jj}{z_{j}}+\mu_{j}})}$}, {$j=1,2$}. The ELBO of VGC-LN is 
\begin{align}
 &\mathcal{L}_{\mathrm{VGC}}(\bd{\mu}, \bd{C})=c_{1}-\mu_{1}+\mu_{2}-\frac{y^2\exp{\left(-\mu_{1}+\frac{C_{11}^2}{2}\right)}}{2}\cr &-\ell_{0}-\exp{\left(\mu_{2}+\frac{C_{21}^{2}+C_{22}^2}{2}\right)}+\ln|\bd{C}|\cr 
 & \ell_{0}={\exp{\left((\mu_{2}-\mu_{1})+\frac{C_{11}^2-2C_{11}C_{21}+C_{21}^{2}+C_{22}^2}{2}\right)}}\nonumber
  \end{align}
where \small{$c_{0}=-0.5\ln{(2\pi)}-2\ln{\Gamma(0.5)}$}, \small{$c_{1}=c_{0}+\ln{(2\pi e)}$}.

The gradients are
\begin{align}
\frac{\partial \mathcal{L}_{\mathrm{VGC}}(\bd{\mu}, \bd{C})}{\partial \mu_{1}} &=-1+{\frac{y^2}{2}\exp{\left(\frac{C_{11}^2}{2}-\mu_{1}\right)}}+  \ell_{0}\cr
\frac{\partial \mathcal{L}_{\mathrm{VGC}}(\bd{\mu}, \bd{C})}{\partial \mu_{2}} &=1-  \ell_{0}-\exp{\left(\mu_{2}+\frac{C_{21}^{2}+C_{22}^2}{2}\right)}
\cr
\frac{\partial \mathcal{L}_{\mathrm{VGC}}(\bd{\mu}, \bd{C})}{\partial C_{11}} &=-\frac{y^2}{2} C_{11}\exp{\left(\frac{C_{11}^2}{2}-\mu_{1}\right)}-(C_{11}-C_{21})  \ell_{0}+\frac{1}{C_{11}}\cr
\frac{\partial \mathcal{L}_{\mathrm{VGC}}(\bd{\mu}, \bd{C})}{\partial C_{21}} &=(C_{11}-C_{21})  \ell_{0}-C_{21}\exp{\left(\mu_{2}+\frac{C_{21}^{2}+C_{22}^2}{2}\right)}
\cr
\frac{\partial \mathcal{L}_{\mathrm{VGC}}(\bd{\mu}, \bd{C})}{\partial C_{22}} &=-C_{22}  \ell_{0}-C_{22}\exp{\left(\mu_{2}+\frac{C_{21}^{2}+C_{22}^2}{2}\right)}+\frac{1}{C_{22}}\nonumber
   \end{align}
   
\subsubsection*{C4:  Stochastic VGC-LN}
 The stochastic part of the ELBO is, 
 \begin{align}
\ell_{s}(\wt{\bd{z}})&=c_{0}+\wt{z_{2}}-\wt{z_{1}}-\frac{y^2\exp(-\wt{z_{1}})}{2}-{\exp(\wt{z_{2}}-\wt{z_{1}})}-\exp(\wt{z_{2}})\nonumber
  \end{align}
and
 \begin{align}
\nabla_{\wt{z}_{1}}\ell_{s}(\wt{\bd{z}})&=-1+\frac{y^2\exp(-\wt{z_{1}})}{2}+{\exp(\wt{z_{2}}-\wt{z_{1}})}\cr
\nabla_{\wt{z}_{2}}\ell_{s}(\wt{\bd{z}})&=1-{\exp(\wt{z_{2}}-\wt{z_{1}})}-\exp(\wt{z_{2}})\nonumber
 \end{align}

\subsubsection*{C5: Stochastic VGC-BP}

\begin{enumerate}
\item   
\small{$\ln p(y, x_{1},x_{2})=c_{0}-2\ln{x_{1}}-{y^2}/{(2 x_{1})}-{ x_{2}}/{ x_{1}}- x_{2}$},  
\begin{align}
 {\partial \ln p(y, x_{1},x_{2})}/{\partial x_{1}}&=-{2}/{x_{1}}+{y^2}/{(2x_{1}^2)}+{x_{2}}/{x_{1}^{2}},\quad \cr
  {\partial \ln p(y, x_{1},x_{2})}/{\partial x_{2}}&=-{1}/{x_{1}}-1\nonumber
 \end{align}
\item  $\Psi(x)$ is predefined as CDF of $\mathrm{Exp}(0.01)$. 
\end{enumerate}

\subsection*{D. Derivations in Poisson Log Linear Regression}
For $i=1,\hdots, n$, the hierarchical model is 
\begin{align}
y_{i}&\sim \mathrm{Poisson}(\mu_{i}), \quad \log(\mu_{i})=\beta_{0} + \beta_{1}x_{i}+\beta_{2}x_{i}^2,\cr  \beta_{0}&\sim N(0, \tau), ~~ \beta_{1}\sim N(0, \tau), ~~ \beta_{2}\sim N(0, \tau),  ~~ \tau\sim \mathrm{Ga}(1,1)\nonumber
\end{align}
The log likelihood and prior,  
\begin{align}
\ln{p(\bd{y}, \bd{\beta}, \tau)}&=\sum_{i=1}^{n} \ln{p(y_{i}|\bd{\beta})}+\ln{\cN(\beta_{0}; 0, \tau)}+\ln{\cN(\beta_{1}; 0, \tau)}\cr &+\ln{\cN(\beta_{2}; 0, \tau)}+\ln{\mathrm{Ga}(\tau; 1, 1)}\nonumber
\end{align} 
where $\ln{p(y_{i}|\bd{\beta})}=y_{i}\ln{\mu_{i}}-\mu_{i}-\ln{y_{i}!}$,  and 
$\mu_{i}=\exp{(\beta_{0} + \beta_{1}x_{i}+\beta_{2}x_{i}^2)}$.

%
%
%
%
%
%
The derivatives are 
\begin{align}
\frac{\partial \ln{p(\bd{y}, \bd{\beta}, \tau)}}{\partial \beta_{0}}=\left[\sum_{i=1}^{n}(y_{i}-\mu_{i})\right]-\tau^{-1}\beta_{0}\nonumber
\end{align}
\begin{align}
\frac{\partial \ln{p(\bd{y}, \bd{\beta}, \tau)}}{\partial \beta_{1}}=\left[\sum_{i=1}^{n}x_{i}(y_{i}-\mu_{i})\right]-\tau^{-1}\beta_{1}\nonumber
\end{align}
\begin{align}
\frac{\partial \ln{p(\bd{y}, \bd{\beta}, \tau)}}{\partial \beta_{2}}=\left[\sum_{i=1}^{n}x_{i}^{2}(y_{i}-\mu_{i})\right]-\tau^{-1}\beta_{2}\nonumber
\end{align}
\begin{align}
\frac{\partial \ln{p(\bd{y}, \bd{\beta}, \tau)}}{\partial \tau}=-\frac{3}{2\tau}+\frac{\beta_{0}^{2}+\beta_{1}^{2}+\beta_{2}^{2}}{2\tau^{2}}+\frac{a_{0}-1}{\tau}-b_{0}\nonumber
\end{align}

\end{document}